%% file: main.tex
\documentclass[pdflatex,iicol,sn-basic,Numbered]{sn-jnl}

\usepackage{graphicx}
\usepackage{booktabs}
\usepackage{array}
\usepackage{multirow}
\usepackage{amsmath,amssymb}
\usepackage{xcolor}
\usepackage{textcomp}
\usepackage{manyfoot}
\usepackage[title]{appendix}
\usepackage{url}
\usepackage[LGR,T1]{fontenc}
\usepackage{tikz}
\usetikzlibrary{positioning,arrows.meta}

\graphicspath{{figures/}}

\newcommand{\gk}[1]{{\fontencoding{LGR}\selectfont #1}}
\newcommand{\gap}{\ensuremath{\square}}
\newif\ifrevmarks \revmarksfalse
\newcommand{\revcolor}{\ifrevmarks\color{red}\fi}
\newcommand{\rev}[1]{\ifrevmarks{\leavevmode\color{red}#1}\else#1\fi}
\newcommand{\revhead}[1]{\texorpdfstring{\ifrevmarks\textcolor{red}{#1}\else#1\fi}{#1}}
\newcommand{\revbibkeys}{angleraud2026structure,vlm2026grounding,todorov2022noise,%
cullhed2024instruct,grammateus2023,idpdata2026,dclp,epidoc,%
preisigke1971supplement,pavlopoulos2024challenging,tmwords,vierros2021papygreek,%
colutto2019transkribus,tsochatzidis2021htr,camps2021abbreviated,strobel2020howmuch,%
strobel2022evaluation,chague2024backgrounds,west2024dating,assael2025aeneas,%
michalcova2026transcription,anagnostes2026,anagnostes2026announcement,%
vanbeek2013quantifying}
\makeatletter
\AtBeginDocument{\ifrevmarks
  \let\rev@bibitem\bibitem
  \renewcommand{\bibitem}[2][]{\normalcolor
    \@for\rev@k:=\revbibkeys\do{\def\rev@t{#2}\ifx\rev@t\rev@k\color{red}\fi}%
    \rev@bibitem[#1]{#2}}%
\fi}
\makeatother

\newcommand{\ESM}{Online Resource~1}
\newcommand{\movednote}[1]{\ifrevmarks{\par\noindent\footnotesize\itshape\color{red}[#1]\par\smallskip}\fi}

\newcommand{\cer}{\ensuremath{c}}
\newcommand{\cninety}{\ensuremath{c_{90}}}
\newcommand{\cninetyfive}{\ensuremath{c_{95}}}

\begin{document}

\title[Which papyrus HTR is good enough?]{Which papyrus HTR is good enough? Character-error-rate tolerance of four papyrological tasks on Greek texts}

\author*[1]{\fnm{Anton} \sur{Repushko}}\email{anton@repushko.com}
\author[2]{\fnm{Elena} \sur{Chepel}}\email{elena.chepel@univie.ac.at}

\affil*[1]{\orgdiv{\orgname{Independent Researcher}, \orgaddress{\city{Berlin}, \country{Germany}}}}
\affil[2]{\orgdiv{\orgname{University of Vienna}, \orgaddress{\city{Vienna}, \country{Austria}}}}

\abstract{%
\textbf{Purpose:} \rev{Most Greek papyri are unpublished, and handwritten text recognition (HTR) could open them to scholars. How accurate it must be for a given papyrological task is unknown. We measure the character error rate (CER) that four such tasks tolerate, with published editions as ground truth.}
\textbf{Methods:} \rev{From the 63,846 Greek texts of papyri.info we derive an error-free, letters-only transcription and degrade it to exact CERs of 1--50\%. We test small models for document type, dating and documentary-versus-literary classification, trained on clean text, and eight search methods (three filters and five ranking methods). We validate the noise on real HTR output: 373 papyri unseen by the Anagnostes recogniser (889 images, 7,584 lines) are compared with synthetic twins degraded to the same CER.}
\textbf{Results:} \rev{The models keep, on average, 90\% of their metric up to 20\% CER for documentary-versus-literary classification, 7.5\% for document type, 5\% for subtypes and search, and 3\% for dating. On real output (median CER 22\%) the tasks lose less than on the twins: document-type classification keeps a macro-F1 of 0.82 against 0.50, and the same is seen in dating. We believe this is because real errors come in runs and leave more text intact; the thresholds are lower bounds.}
\textbf{Conclusion:} \rev{The study gives a CER target for each task, releases corpora and code, and shows that imperfect recognition is already useful for some tasks.}%
}

\keywords{Ancient Greek papyri, handwritten text recognition, character error rate, noise robustness, text classification, information retrieval}

\maketitle

\section{Introduction}\label{sec:intro}

Papyri are the largest body of first-hand written evidence for the ancient Mediterranean, and most of them have never been \rev{transcribed}. Large collections worldwide hold in total between 1,000,000 and 1,500,000 excavated fragments, of which less than 10\% has been edited and published despite more than a century of papyrological work~\cite{vanminnen2009future}. Each edition requires \rev{weeks and often months} of expert labour: the papyrologist reads a damaged, often cursive text written without word division, restores lost passages, expands abbreviations and normalises spellings\rev{. Transcription and interpretation are not separate stages of this work: to restore a lost passage the editor must already compare the text with other documents, identify its type and date it. But only the finished edition makes the full content of the papyrus available to other scholars, to historical analysis and to the databases.} \rev{The text of the} unedited majority of papyri \rev{remains} invisible to digital tools, and a scholar who wants to know whether a formula, a name or a certain \rev{term} occurs among the unpublished papyri has no way to ask. \rev{Online catalogues of collections often record the type of an unpublished papyrus, but its full text is not available.}

Automatic transcription would change this. If handwritten text recognition (HTR) could turn images of unedited papyri into text, even though imperfect, this text layer could be searched for parallels, sorted by type and date, before each single fragment receives a proper papyrological edition, and the \rev{limited} expert attention could \rev{be directed to the most promising fragments}. In the last few years, the ICDAR~2023 competition on detecting and recognising Greek letters on papyri~\cite{seuret2023icdar} and its follow-ups~\cite{turnbull2024papyri}, citizen-science character datasets~\cite{swindall2021citizen}, end-to-end systems such as Anagnostes~\cite{anagnostes2026}, and, for the printed side, OCR of Greek critical editions~\cite{robertson2017greekocr,angleraud2026structure,vlm2026grounding} have all shown some progress in this direction.

In these previous studies, character error rates have been reported as performance metrics. However, the authors have not answered the question which would be relevant for a papyrologist or a collection curator: \emph{which} of these error rates are \rev{low} enough for \emph{which} tasks? Studies on modern and early-modern print have shown that the impact of OCR noise is strongly task dependent~\cite{traub2015impact,chiron2017impact,hill2019dirty,vanstrien2020impact,hamdi2023indepth,todorov2022noise}, but none of them covers \rev{the conditions that papyri combine}: scriptio continua without word division, lacunae inside the text, fragments of a few letters\rev{, abbreviations and special symbols, and} a highly formulaic documentary language\rev{. Some of these features are shared with other manuscript traditions (Sect.~\ref{sec:related}), but the effect of recognition errors on downstream tasks has not been measured for such texts}.

In this paper, we directly measure tolerance. We start with the text of \rev{the editions in papyri.info} and work backwards: stripping each edition of everything an HTR system cannot see\rev{, which leaves what a recogniser would return if it read every preserved letter correctly and did nothing more (an \emph{error-free letters-only transcription})}; then degrading that transcription to various CERs and observing their effect on each of the four papyrological tasks. \rev{Because the errors are synthetic, we then check them against real ones: 373 papyri (889 images, 7,584 lines of text) are transcribed by the Anagnostes system, and each transcription is compared with synthetic twins degraded to the same CER (Sect.~\ref{sec:real}).}

\rev{The contributions are of four kinds:
\begin{enumerate}
\item \textbf{A tolerance study.} For four papyrological tasks with real labels we measure how models trained on clean text behave at CERs of 1--50\%, by fragment size, with whole lines lost and with the errors distributed in four different patterns (substitutions only, a uniform mix, bursts, uneven lines), and report a CER threshold per task and model (Sects.~\ref{sec:protocol}--\ref{sec:cross}).
\item \textbf{A validation on real HTR output.} On papyri that the recogniser had never seen, real output is compared with synthetic noise of the same number of errors (Sect.~\ref{sec:real}).
\item \textbf{Released resources.} The rules that turn an EpiDoc edition into an error-free letters-only transcription, the degradation algorithm with its verified Levenshtein invariant, the 576 frozen corpora, the task labels (those of Grammateus are joined from the project's own export), splits, models and result tables (Sects.~\ref{sec:data}--\ref{sec:models}).
\item \textbf{Recommendations} for HTR developers, tool builders and papyrologists: what to report alongside a CER, and which uses the error rates of current systems support (Sect.~\ref{sec:discussion}).
\end{enumerate}
There are two main findings. Tolerance differs between tasks by close to an order of magnitude. Real HTR output damages the tasks less than synthetic noise with the same number of errors, so the thresholds are lower bounds for real output. An error-free transcription is not necessary: for classification and search the required CER lies within reach of current systems, at least on clearly written hands, while dating from the text alone is limited by the model as much as by the recognition.}

\rev{Supporting detail is in \ESM: its sections, figures and tables are cited with the prefix S (Sect.~S1, Fig.~S1, Table~S1).}

\section{Related work}\label{sec:related}

\paragraph{HTR for Greek papyri and OCR for Greek print}
Character-level detection and recognition on papyrus images was the objective of the ICDAR~2023 competition~\cite{seuret2023icdar}, whose winning recognition entry combined YOLOv8 detection with DeiT and SimCLR classifiers~\cite{turnbull2024papyri}. Earlier work used crowdsourced character annotations from the Ancient Lives project to train and evaluate classifiers~\cite{swindall2021citizen}. Line- and page-level transcription systems are more recent; Anagnostes~\cite{anagnostes2026} \rev{reports an average letters-only CER of 17.9\% on a held-out validation set of 914 lines from 55 papyri, across all types of handwriting: from 12\% on clearly written lines to 26\% on lines with lacunae, with the best results between 1.9 and 8.6\%~\cite{anagnostes2026announcement}}. For printed polytonic Greek, large-scale OCR of critical editions reaches low CERs on clean nineteenth-century typography~\cite{robertson2017greekocr}, and recent transformer systems handle the structure of critical editions~\cite{angleraud2026structure}; vision-language models, by contrast, have been shown to guess rather than read Greek~\cite{vlm2026grounding}. None of these works measures what their output is useful for.

\paragraph{\revhead{HTR for historical manuscripts and post-correction}}
\rev{Between printed Greek and papyri lies the recognition of medieval and early-modern manuscripts, which share much with papyri: abbreviations, irregular word division, damage and little training material per hand. Platforms such as Transkribus let scholars train their own models and make their documents searchable by keyword spotting~\cite{colutto2019transkribus}, and for Greek historical handwriting a dedicated recogniser has been evaluated on three new collections~\cite{tsochatzidis2021htr}. Much of this work asks what a recogniser should be trained on and how its output is to be judged: whether abbreviations are expanded by the HTR engine itself or by a separate normalisation step~\cite{camps2021abbreviated}, how much ground truth a given accuracy requires~\cite{strobel2020howmuch}, whether ground truth augmented with artificial backgrounds serves historical pages~\cite{chague2024backgrounds}, and how a model can be evaluated when no ground truth exists~\cite{strobel2022evaluation}. Where this literature speaks of sufficient quality it does so in general bands of character accuracy, above 98\% ``good'' and below 90\% ``poor''~\cite{strobel2022evaluation}, while noting that for indexing and text mining a word-level measure gives a better picture than the CER~\cite{strobel2020howmuch}. We measure instead what a given CER does to particular tasks, and the bands do not fit them: one of our tasks works at 20\% CER and another needs 3\%. As for the errors themselves, Pavlopoulos et al.\ show that correcting recognised Byzantine Greek automatically remains a hard, only partly solved problem~\cite{pavlopoulos2024challenging}. Post-correction repairs the transcription; we ask how much damage the tasks tolerate without it.}

\paragraph{Computational papyrology and epigraphy on transcribed text}
Most machine-learning work on ancient Greek text takes edited text as input. Ithaca restores lacunae and attributes inscriptions to a place and a date~\cite{assael2022ithaca}\rev{.} Several approaches have been applied to dating papyri. Text regression estimates a document's date \rev{based on its text}~\cite{pavlopoulos2023dating}, while instruction-tuned language models restore, date and localise both papyri and inscriptions~\cite{cullhed2024instruct}. The Hell-Date benchmark, in turn, compares image-based dating methods with the judgements of palaeographers~\cite{degregorio2024helldate}. All of the text-based systems assume clean edited input; our question is what happens to such tasks when the input comes from HTR instead.

\paragraph{Impact of OCR noise on downstream tasks}
For printed historical documents the effect of OCR noise on research tasks has been studied repeatedly. Traub et al.~\cite{traub2015impact} and Chiron et al.~\cite{chiron2017impact} looked at retrieval and use in digital libraries, Hill and Hengchen~\cite{hill2019dirty} at text analysis on eighteenth-century print, van Strien et al.~\cite{vanstrien2020impact} across a range of NLP tasks, Hamdi et al.~\cite{hamdi2023indepth} at named-entity recognition and linking, and Todorov and Colavizza~\cite{todorov2022noise} at language models; post-OCR correction is surveyed in~\cite{nguyen2021survey} \rev{and taken up for recognised Byzantine Greek in~\cite{pavlopoulos2024challenging}}. These studies share our conclusion that impact is task dependent, but they work on running text of substantial length with word divisions and mostly with real OCR output whose error rate cannot be varied at will. We instead control the error rate exactly, vary its shape, and evaluate tasks and text that are specific to papyrology.

\section{Data}\label{sec:data}

\subsection{Sources}

\textbf{papyri.info.} The Integrating Digital Papyrology data (idp.data) holds the EpiDoc XML encoded editions of the Duke Databank of Documentary Papyri (DDbDP), the Digital Corpus of Literary Papyri (DCLP) and the metadata of the Heidelberger Gesamtverzeichnis (HGV)~\cite{idpdata2026,hgv,dclp,epidoc}. We pin one commit (\texttt{fd88c0c}, \rev{committed on 6 July 2026 and accessed on} 13 September 2026) and take every \rev{text} whose first \texttt{div[@type='edition']} is in Ancient Greek \rev{as it stands at that commit (``the edition'' of a papyrus here means its text as it existed on 6 July 2026)}: 66,591 DDbDP and 6,032 DCLP files. HGV supplies \rev{metadata only: we use the date interval it records for a documentary text, while the text itself always comes from DDbDP. HGV does not cover the literary papyri of DCLP.} The Trismegistos number (TM)~\cite{depauw2014trismegistos} identifies the physical papyrus across databases.

\textbf{Grammateus.} The Grammateus project classifies documentary papyri into four types and 25 subtypes\rev{ (24 distinct ones, since receipts appear under two types)} by their formal structure~\cite{grammateus2023}. Its export of 13 September 2026 \rev{has 1,843 entries for 1,840 papyri; 1,780 match a text of our corpus by its identifier and 1,776 papyri carry a type (most of the rest are cited under another identifier and were not matched).}

\textbf{Preisigke's W\"orterbuch.} \rev{The W\"orterbuch der griechischen Papyrusurkunden, begun by Friedrich Preisigke and continued in supplements, is the standard dictionary of the vocabulary of Greek documentary papyri.} The OCR text of \rev{its} Supplements 1--3~\cite{preisigke1971supplement} on the Internet Archive supplies dictionary headwords from which search-query stems are drawn (Sect.~\ref{sec:t4data})\rev{; the main volumes are available there as well, but their OCR text contains no Greek characters, so only the supplements could be used}.

\subsection{Corpus}

Every edition is \rev{reduced to its letters-only text (L2, Sect.~\ref{sec:cer})}. DDbDP records whose text is reprinted in another record (8,777 \texttt{reprint-in} stubs) are dropped, leaving 63,846 documents with 18.6 million letters: 57,814 documentary (DDbDP) and 6,032 literary or subliterary (DCLP). Letters per document are heavily skewed: 10th percentile~9, median~106, 90th percentile~632, maximum~161,212. 4,547 documents have no letters at all, 3,860 of them DCLP catalogue records without a transcription; they cannot be predicted from text and are excluded from the model metrics.

\subsection{Splits}\label{sec:splits}

Documents are grouped so that no papyrus and no near-duplicate text can appear on both sides of a split: two texts land in one group if they belong to the same document (established by a shared TM number) \rev{or} if their letters-only 5-gram sets have an estimated Jaccard similarity of at least 0.8 (MinHash with 128 permutations~\cite{broder1997resemblance})\rev{; the groups are the connected components of these two relations, so either condition is enough}. Groups are assigned to train, validation and test (80/10/10) by a seeded hash of the group's smallest identifier, giving 51,042, 6,370 and 6,434 documents. \rev{Dating, documentary versus literary and search all use this one split, each restricted to the documents that carry its label (Table~\ref{tab:tasks}), so they are trained and tested on overlapping documents.} The document-type task has too few labelled papyri for a fixed split and uses five-fold grouped cross-validation stratified by type (folds of 355--356 papyri). Hyperparameters of every model are tuned once on clean validation text \rev{(the validation split or, for document type, a fold held out from training)} and frozen for every noise level.

\section{Tasks and the data behind each of them}\label{sec:tasks}

Table~\ref{tab:tasks} lists the four tasks. Each has a headline metric and a baseline (the majority class or the median date). Tasks were chosen to cover what a papyrologist would \rev{first} want to do with an unedited text\rev{, at the stage of cataloguing}: \rev{identify what type of document it is (T1), date it (T2), tell a document from a literary fragment (T3), and establish its thematic content with keywords present in the text (T4)}. Literary attribution and editorial text reconstruction are left for later work\rev{, as are other uses of HTR output, such as aligning image and text in a digital edition or assisting the editorial transcription itself (word division, resolution of abbreviations)}.

\begin{table*}[!t]
\caption{The four tasks, the data behind each and how they are scored. Counts are documents with at least one letter; validation and test are the evaluated subsets. \rev{The sets of T2--T4 are not separate splits: each is the part of the single corpus split of Sect.~\ref{sec:splits} that carries the task's label.} T1 uses five-fold grouped cross-validation over all its papyri}
\label{tab:tasks}
\centering\footnotesize\setlength{\tabcolsep}{4pt}
\begin{tabular}{@{}p{2.2cm}p{2.5cm}p{3.1cm}p{2.5cm}p{2.3cm}p{1.4cm}@{}}
\toprule
Task & Labels from & Train / validation / test & Target & Headline metric & Baseline \\
\midrule
T1 Document \rev{type} & Grammateus type & 1,776 (5-fold CV) & 4 types & macro-F1 & 0.121 \\
T1 subtype & Grammateus subtype & 1,665 (5-fold CV) & 16 subtypes ($\geq$30 papyri) & macro-F1 & 0.028 \\
T2 Dating & HGV date interval & 43,803 / \rev{4,591} / 4,681 & year (interval midpoint) & MAE in years & 184 \\
T3 Documentary/literary & DDbDP vs DCLP & 47,328 / 5,906 / 5,950 & 2 classes & macro-F1 & 0.491 \\
T4 Search & dictionary, names, formulae, substrings & 2,534 queries; 5,961 docs, 96,410 lines & relevant units & recall@20, filter F1 & -- \\
\bottomrule
\end{tabular}
\end{table*}

\subsection{T1: document type}
Papyrologists identify the genre of the document by \rev{formulae}, as well as by the structure and layout of the text. For this task, both for humans and machines, the size of the fragment and the amount of preserved words are crucial. In the Grammateus dataset that we used, the majority of papyri are complete or near-complete.

The Grammateus typology groups documentary papyri by the formal relation they establish between the parties: \emph{Epistolary Exchange} (569 papyri in our set), \emph{Objective Statement} (483), \emph{Transmission of Information} (437) and \emph{Recording of Information} (287). Below the types are subtypes such as receipt, list, business letter, syngraphe, private letter, cheirographon or petition; we keep the 16 subtypes with at least 30 papyri (1,665 documents) as a harder 16-class variant. \rev{Receipts, which Grammateus lists under two types, form one class here.} The metric is macro-F1 over the pooled folds. The labelled papyri are long (median 410 letters), so this task sees more text per document than the others.

\rev{Grammateus is a curated selection, not an exhaustive classification: it describes 1,840 papyri, against the roughly 58,000 documentary texts of our corpus, and an unedited papyrus may belong to none of its types, or not be a document at all. Our classifiers always assign one of the four types. A tool used in practice would also need the answer ``unknown'' for papyri that fit none of them. We did not test this, so the T1 thresholds hold for the easier case in which every papyrus belongs to one of the four types.}

\subsection{T2: dating}
Papyrologists usually establish the date of a document by looking at the handwriting and searching the text for historical indications, such as the name of a known person, economic or social institutions, and others. If the text of the fragment contains no dating formula or other restrictive indications, this dating remains subjective. By convention, such rough dating is placed within the boundaries of a century, as reflected in the HGV metadata. \rev{A model trained on these dates uses only the text, but it does not escape the convention: its targets are the editors' intervals, so it learns to reproduce their judgement, century boundaries included, and not an independent ground truth.}

HGV gives a date interval for 55,410 of our documents\rev{, all but three of them documentary (none of the three is in the test set): HGV supplies the label and nothing else (the text is the DDbDP transcription), and the literary papyri of DCLP are not part of this task. The models take no feature from HGV: their only input is the letters-only text (letter $n$-grams for TF-IDF and fastText, the letter sequence for the neural models). A text-only model dates the composition of a text, not the writing of the object, so a first-century copy of Homer would be dated by Homer's language; for documents the two coincide, which is why the task is restricted to them}. Training uses every dated document (43,803) with the interval midpoint as target; evaluation is restricted to documents whose interval is at most 100 years wide (4,681 evaluated test documents), so that the reference date is meaningful. The headline metric is the mean absolute error (MAE) in years; we also record the share of predictions inside the editor's interval and the century accuracy. The always-median baseline errs by 184 years. Dates range from the fourth century BCE to the \rev{ninth} century CE with a peak in the second century CE \rev{(Table~\ref{tab:centuries}); the set is far from balanced, with 29\% of the evaluated documents in the second century CE. This reflects the uneven distribution of the surviving papyri over the centuries, which peaks in the second century CE~\cite{vanbeek2013quantifying}.}

\subsection{T3: documentary versus literary}
Documentary and literary papyri constitute two very different groups of texts that are studied by historians and philologists using different approaches. The two groups already differed greatly in quantity in antiquity: the vast number of everyday documents far exceeded the number of literary works copied on papyrus. Text loss is also distributed unevenly between documentary and literary fragments. Literary fragments usually represent only tiny portions of a long poem or book, whereas documentary papyri often preserve complete or nearly complete documents. From the lexical point of view, literary and documentary papyri often have very little in common, since documents reflect bureaucratic, legal, and mundane language, while literature tends to use sublime, elevated, and archaic style.

A text from DDbDP counts as documentary and a text from DCLP as literary or subliterary. \rev{DCLP is not limited to literature in the narrow sense: it also holds paraliterary texts, such as school exercises and magical, medical and astrological texts, and more generally what was not taken into DDbDP. Their language can be close to that of documents, which adds noise to the boundary between the two classes; the ``literary'' class of this task is literary in this broad sense.} Papyri present in both databases (by TM) are removed. Only 3.4\% of evaluated test documents are literary, because most DCLP records carry no transcription, so macro-F1 carries the headline. To check that the task is not solved by the number of lines and the number of letters alone, a model that sees only line-length statistics is reported alongside the text models.

\subsection{T4: search}\label{sec:t4data}
The first examination of an unedited papyrus by a papyrologist, whether for catalogue description or the preparation of the first transcription, involves identifying words and phrases that are relevant to the document\rev{'}s content and may point to parallels in other papyri. Search is the task most directly tied to unedited material: a papyrologist would want to look for a word, a name or a formula among unpublished papyri in a collection. Queries are letters-only strings of four kinds: 1,000 stems of Preisigke headwords (the headword with its longest inflectional ending removed), 500 capitalised word forms from the editions (proper nouns: people, places, gods, months), 34 documentary formulae such as \gk{qairein}, \gk{etous} or \gk{errwsjai se euqomai}\rev{, chosen by the authors,} and 1,000 random substrings of 3--12 letters of clean test lines. \rev{Long queries are not at a disadvantage: only 34 of the 2,534 queries are formulae, and the longer a query is, the more noise the search tolerates (Sect.~\ref{sec:results}).} \rev{\ESM{} (Sect.~\ref{esm:queries}) describes how each set is drawn and lists the formulae.} \rev{The Preisigke stems stand for dictionary look-up, a search for a lexeme whatever its inflection. Queries and texts carry no word spaces, because the papyrus has none: a formula of several words is matched as one string of letters (Sect.~\ref{esm:queries} discusses lemmatised sources and word-divided text, which we did not use).}

A unit (a document or a line) is relevant to a query when the query occurs in its \emph{clean} text as an exact substring; relevance never changes with noise. \rev{A papyrus or one of its lines thus counts as a correct answer only if its clean text contains the query letter for letter, whichever method is tested (Sect.~\ref{sec:models}). This is strict on purpose: when an approximate method finds a spelling variant of the query, or another word one letter away from it, this is counted as a false hit, although a papyrologist might welcome it, so the precision of such methods is low even on clean text. The comparison across CERs is unaffected, because the truth is the same at every CER.} There are 5,961 test documents and 96,410 test lines with letters. Ranking methods are scored by recall@20 (relevant units in the top 20 divided by $\min(\text{relevant}, 20)$), MRR and nDCG@10; filtering methods by micro precision, recall and F1 over all hits.

\section{From the edited text to HTR output at an exact CER}\label{sec:cer}

\begin{figure*}[!t]
\centering
\input{figures/fig1.tex}
\caption{A real papyrus from the edition to letters-only text and its degradation. Lines \rev{19--20} of the letter of Apion to his father (BGU II 423 = Chrest.Wilck.\ 480) as encoded in the DDbDP edition (L0; square brackets enclose restored letters, dots mark uncertain ones), with the editorial layer removed (\rev{L1, preserved text}; \gap{} marks lost text), and reduced to letters (\rev{L2}), the text \rev{an error-free} HTR system would return. The last three \rev{blocks} are the same lines degraded by the study's algorithm to 5, 15 and 30\% CER \rev{(the number of edits is whole)} with the default 3:1:1 mix: a substituted letter is bold, an inserted letter bold and underlined, and a deleted letter is marked by a bold dot. Every edit costs one Levenshtein step and gap tokens are never touched\rev{. Fig.~\ref{fig:pipelinefull} adds lines 1--2}}
\label{fig:pipeline}
\end{figure*}

Figure~\ref{fig:pipeline} shows the three levels and their degradation on a real papyrus. The first half of the pipeline answers a question that has no obvious answer: what \rev{should count as the error-free HTR output for} a papyrus? An edition is not it. Editors restore lost letters in brackets, resolve abbreviations, normalise errors and omissions, add accents, breathings and word spaces, and mark uncertain readings; none of this is on the papyrus, and an HTR system that returned it would be doing philology, not recognition. We therefore define three levels and measure only the last. \rev{The target is deliberately narrow: the letters preserved on the papyrus. What an HTR system should output depends on the editorial tradition it serves~\cite{michalcova2026transcription}, and a system that also expands abbreviations, or records the abbreviation marks that our rules drop, does more than this target asks; its output has to be reduced to the letters before our thresholds apply.}

\subsection{Levels}

\textbf{L0, the edition}, is the EpiDoc XML encoded text. \textbf{\rev{L1, the preserved text,}} removes the editorial layer by the rules listed in \rev{Sect.}~\ref{app:ink}: whatever the scribe wrote stays, whatever the editor added goes, and lost text becomes one gap token \gap{} per contiguous loss. \textbf{\rev{L2}, letters only}, normalises what remains: Unicode NFD, combining marks dropped, lowercase, the \rev{sigma forms} folded to one, and only the 24 letters $\alpha$--$\omega$ plus the three archaic numeral letters stigma, koppa and sampi kept; accents, breathings, punctuation, brackets, digits and other scripts are removed. Line breaks are kept, because lines are visible on the papyrus and HTR systems work line by line; word spaces are dropped, because the papyrus has none. \rev{(The released code and corpora call the letters-only level L3.) Lower case and a single sigma are choices of encoding, not a loss of information (Sect.~\ref{app:ink}). Only L2 is degraded and scored. L0 and L1 still carry the editor's accents, punctuation, capitals and word division, which are not on the papyrus and which a recogniser cannot be asked to reproduce.}

The rules were fixed against editions before any experiment and are documented in the released code; the element inventory that motivated them (Table~\ref{tab:ink}) was measured over \rev{all 72,623 Greek editions}. Two decisions deserve comment. Restorations are removed even though a strong language model could guess many of them, because an HTR system sees no ink there; lacuna restoration is a separate task. Uncertain letters (\texttt{unclear}, the dotted letters of an edition) are kept, because the ink is there and a recognition system would produce \emph{something}; whether it produces the right letter is exactly what the degradation models. \rev{The opposite assumption, that a recogniser gets nothing from a doubtful letter, would turn every \texttt{unclear} element into a gap; we did not run the experiments under it. Real errors can be expected to fall on these letters more often than on others (Anagnostes reports 12\% CER on clearly written lines against 21\% on abraded ones~\cite{anagnostes2026announcement}), which the even placement of the default noise does not reproduce.}

\subsection{Structure tokens and CER}\label{sec:cerdef}

The gap token and the line break are structure, not letters. They are never degraded and never scored, and they stop search hits from spanning a lacuna, as they should. \rev{The extent of a loss is not encoded: a gap of one letter and a gap of half a line are the same token. The document-level models see the token as one more symbol of the input string, so the number and position of the gaps and the division into lines are available to them as features, but not the size of the gaps. The line-length control of T3 (Sect.~\ref{sec:models}) bounds what layout alone can give; what the gap tokens themselves contribute to the document-level tasks was not measured.} CER is the letters-only Levenshtein distance~\cite{levenshtein1966binary} between reference and hypothesis divided by the number of reference letters, micro-averaged over the evaluated text:
\begin{equation}
\mathrm{CER} = \frac{\sum_d \mathrm{Lev}(\mathrm{ref}_d, \mathrm{hyp}_d)}{\sum_d |\mathrm{ref}_d|},
\end{equation}
where the sums run over documents $d$ and the strings contain letters only. This is the quantity an HTR developer should report to make use of our thresholds: computed on letters after the same normalisation, without diacritics, case, punctuation, spaces or bracketed restorations.

\subsection{Degradation at an exact CER}\label{sec:degrade}

Real HTR output \rev{with aligned ground truth does not exist at the scale this study needs, tens of thousands of documents to train and to test on,} and its error rate cannot be set. We therefore degrade the \rev{L2} text synthetically\rev{, and compare the synthetic errors with real HTR output on a subset of papyri in Sect.~\ref{sec:real}. The algorithm makes the CER exact and the errors well defined. For a split (train, validation or test) and a target CER \cer{}, $K = \mathrm{round}(\cer \cdot N)$ edits, $N$ being the letters of the split, are apportioned to the documents in proportion to their letters, so the CER of the split equals \cer{} exactly. The edits of a document fall on random, pairwise non-adjacent letters and are substitutions, insertions and deletions in the proportion 3:1:1; gap tokens are never edited. The Levenshtein distance between the clean and the degraded letters of every document must equal its number of edits, so that every edit costs exactly one step and none is cancelled by another: the distance is computed for every document of every corpus, a line on which it falls short is redrawn, and no document in any of the 576 corpora violates the invariant. Every edit is logged and every random draw is seeded, so a truth defined on the clean text can be projected onto the degraded text and each document is reproducible. \ESM{} (Sect.~\ref{esm:degrade}) gives the rules in full.

\textbf{Short documents.} Edits are whole, so a short fragment is either untouched or noisier than nominal: at a nominal 5\% CER a fragment of 12 letters receives one edit (8.3\%) and a fragment of 8 letters none. From 100 letters upward the CER a document receives equals the nominal one to within 0.1 points, but no document under 20 letters is edited at 1\% or 2\% CER (Table~\ref{tab:effcer}). Thresholds at or below 3\% CER, where dating sits, are therefore carried by documents of about 50 letters or more.}

\rev{\textbf{Error-shape variants.}\label{sec:variants} Recognition errors are not single letters scattered evenly over a text. Four variants spend the same edit budget in a different pattern, on the test corpora of T1, T3 and T4 (dating was not run under them): (i) \emph{substitutions only} (1:0:0); (ii) a \emph{uniform mix} (1:1:1); (iii) \emph{bursts}, in which the substitutions of the default mix are grouped into runs of 2--4 consecutive letters, as when a recogniser loses a damaged word; (iv) \emph{uneven lines}, in which the edits of a document are spread over its lines with log-normal weights ($\sigma = 0.5$), so that some lines are much noisier than others. In every variant the corpus CER is still exactly \cer{} and the invariant holds (results in Sect.~\ref{sec:shape}).}

\textbf{Lost lines.} Independently of letter noise, a seeded fraction $p \in \{10, 20, 30\}\%$ of a document's text-bearing lines is dropped, crossed with $\cer \in \{0, 5, 10, 15, 20, 30\}\%$. CER is measured on the lines that remain, so the two kinds of loss stay separate: a line the layout analysis missed is not a letter the recogniser got wrong.

\textbf{Grid and seeds.} The CER grid is \cer{} = 0, 1, 2, 3, 5, 7.5, 10, 12.5, 15, 17.5, 20, 25, 30, 40 and 50\%. Test corpora are degraded with three noise seeds\rev{. Degraded training and validation corpora, with one seed each that is never shared with the test text, are part of the release but are not used in this paper}; the 576 frozen corpora are released with the code.

\section{Models}\label{sec:models}

The models are deliberately small and trained in this study (Table~\ref{tab:models}); no external pretrained system other than the ByT5-small checkpoint is used. For each document-level task (T1--T3) we use one family from each of four kinds: a linear model on character $n$-grams, a shallow embedding classifier, a character convolutional network and a pretrained byte-level transformer. Search uses eight untrained string and term-weighting methods.

\begin{table*}[!t]
\caption{Models\rev{, in brief; architectures and frozen settings are in \ESM{} (Sect.~\ref{app:settings})}. All hyperparameters are tuned once on the clean validation split and frozen}
\label{tab:models}
\centering\footnotesize\setlength{\tabcolsep}{4pt}
\begin{tabular}{@{}llp{11.4cm}@{}}
\toprule
Model & Tasks & Description \\
\midrule
TF-IDF + logistic / ridge & T1--T3 & \rev{letter 1--5-gram TF-IDF with logistic regression (classes) or ridge regression (dates)}~\cite{pedregosa2011sklearn}\\
fastText & T1--T3 & \rev{supervised fastText on letter 3--5-grams}~\cite{joulin2017fasttext}\rev{; dating as 25-year classes}\\
Char-CNN & T1--T3 & \rev{residual character convolutional network, 2.6M parameters}~\cite{zhang2015charcnn}\\
ByT5-small & T1--T3 & \rev{byte-level encoder of ByT5-small, 218M parameters, with a linear head}~\cite{xue2022byt5}\\
Line-length shortcut & T3 & gradient-boosted trees on line-length statistics only: does layout give the answer away?\\
Search methods & T4 & \rev{three filters: exact match and Levenshtein distance of at most 1 or 2 edits; five ranking methods: normalised distance (hits within $0.25$, ranked), minimum edit distance}~\cite{myers1999bitvector}\rev{, trigram overlap, TF-IDF cosine and BM25}~\cite{robertson2009bm25}\rev{; all of these but minimum edit distance also give a filter}\\
\bottomrule
\end{tabular}
\end{table*}

\rev{\textbf{Training.} Every model is trained on clean L2 text and tested at every CER, which is what happens when a tool built on editions is pointed at HTR output.}

\section{Experimental protocol}\label{sec:protocol}

\subsection{Curves and intervals}
Every task $\times$ model gives a curve $M(\cer)$ of the headline metric against CER. Its uncertainty is a percentile bootstrap~\cite{efron1993bootstrap} with 1,000 draws that resamples the evaluation units (documents, or queries for T4) and the noise seeds, using the same draws at every CER so that neighbouring points are paired. Metrics pool the seeds' predictions rather than averaging per-seed scores. Model training is not resampled, so the intervals do not cover \rev{the variation between training runs}.

\subsection{Retention and thresholds}\label{sec:thresholds}
Tasks live on different scales, so each curve is also expressed as \emph{retention}, the share of clean performance kept: $M(\cer)/M(0)$ for higher-is-better metrics and $M(0)/M(\cer)$ for MAE, with a paired interval from the same draws, so that document-to-document variation shared by \cer{} and 0 cancels. From the retention curve we read two thresholds: \cninetyfive{} (\cninety{}) is the highest CER on the grid such that, at every level up to it, the 95\% lower bound of retention is at least 0.95 (0.90). The contiguity requirement and the use of the lower bound make the thresholds conservative.

\rev{\textbf{Task-level thresholds.} A statement about a task, as opposed to a model, needs one curve per task. We average the retention curves of the task's models, each weighing the same and all scored on the same bootstrap draw, and read \cninetyfive{} and \cninety{} from the mean curve: over the four trained models for T1--T3 (three for the subtype variant; the line-length control is left out) and over the five ranking methods at document level for T4. The abstract quotes these task-level, or model-averaged, thresholds; Table~\ref{tab:targets} sets them next to that of the best single model.}

\subsection{Fragment size}
Papyri are mostly small, and an average over documents is an average between fragments with 20 letters and papyrus scrolls containing 20,000 letters. Every curve is therefore also computed per size bin: letters per document (1--19, 20--49, 50--99, 100--199, 200--499, 500--999, 1,000--1,999, $\geq$2,000) and text-bearing lines for T1--T3, letters per query for T4. Bins with fewer than 30 units get no threshold. \rev{In the shortest bins the nominal CER is quantised (Sect.~\ref{sec:degrade}, Table~\ref{tab:effcer}).}

\section{Results by task}\label{sec:results}

Each task has a figure with the headline curves of every model, trained on clean text; \rev{Sect.}~\ref{app:thresholds} lists the thresholds per model. Numbers in the text are point estimates on the test split \rev{(the pooled folds for document type)}; \rev{their 95\% intervals are in the released tables and are not drawn in the figures; their half-width is at most 0.026 macro-F1 for document type, 6.5 years for dating, 0.031 macro-F1 for documentary versus literary and 0.014 recall@20 for search.} \rev{The curves show synthetic noise; real HTR output is set against it in Sect.~\ref{sec:real}, on its own set of papyri.}

\subsection{T1: document type}
\begin{figure}[!ht]
\centering
\includegraphics[width=\columnwidth]{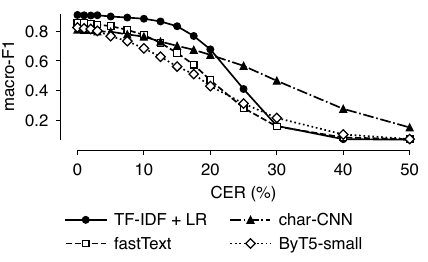}
\caption{T1 document type: macro-F1 of every model against CER}
\label{fig:t1}
\end{figure}

On clean text the linear model is clearly best \rev{(Fig.~\ref{fig:t1})}: TF-IDF with logistic regression reaches macro-F1 0.909, with fastText, ByT5-small and the char-CNN between 0.811 and 0.857 (the neural models train on three of the five folds and stop early on a fourth, which accounts for part of the gap). Under noise the ranking inverts. \rev{TF-IDF} loses little up to 10\% CER and keeps 90\% of its clean score to 12.5\% (\rev{Sect.}~\ref{app:thresholds}\rev{; the task-level threshold, averaged over the four models, is 7.5\%}), then collapses to 0.163 at 30\%, barely above the 0.121 majority-class baseline: the letter $n$-grams it matches on stop occurring. The char-CNN, which scores local context rather than exact $n$-grams, degrades gradually instead and is the best model \rev{from 25\% CER up} (0.466 at 30\%). At a fixed error rate the arrangement of the errors matters as much as their number (\rev{Sect.~\ref{sec:shape}}), and losing 30\% of the text-bearing lines costs \rev{about as much as 10\% CER does (\rev{Sect.}~\ref{app:lost})}. \rev{The 16-subtype variant is harder and less tolerant: its task-level threshold is 5\%, the figure quoted in the abstract, and that of its best model 10\% (Sect.~\ref{app:thresholds}).}

\subsection{T2: dating}
\begin{figure}[!ht]
\centering
\includegraphics[width=\columnwidth]{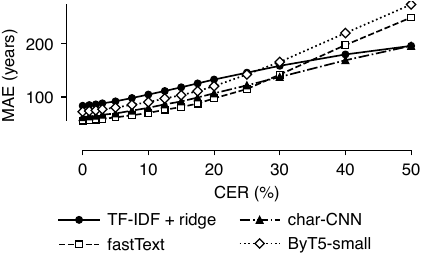}
\caption{T2 dating: mean absolute error in years (lower is better) of every model against CER}
\label{fig:t2}
\end{figure}

Dating is the most fragile task in the study, and the only one where every model degrades at the same rate \rev{(Fig.~\ref{fig:t2})}. On clean text fastText, predicting a probability-weighted midpoint over 25-year bins, dates test documents with a mean absolute error of 55 years, against 62 for the char-CNN, 72 for ByT5-small, 83 for ridge regression and 184 for the median-date baseline. Every model loses about a tenth of its accuracy by 5\% CER and a fifth by 10\%, so even the best keeps 90\% of its clean accuracy only up to 3\% CER (\rev{Sect.}~\ref{app:thresholds}); at 20\% the error is 97 years for fastText, and at 30\% the char-CNN takes over.

\rev{The two other metrics tell the same story. On clean text fastText puts 67.6\% of the test documents in the right century (always answering with the most frequent century would be right for about 29\%) and 19.1\% inside the editor's interval (char-CNN 54.8\% and 13.1\%, ByT5-small 50.7\% and 11.6\%, ridge regression 41.4\% and 10.3\%); at 10, 20 and 30\% CER its century accuracy is 60.7, 49.2 and 33.5\% and its in-interval share 15.8, 12.2 and 8.4\%. The in-interval share is low by construction, since 43\% of the evaluated documents are dated to an exact year.} No model reaches the century-level answer papyrologists ask for, even on \rev{error-free text, where the best one misses the century of a third of the documents}, so for dating the modelling question comes before the recognition one; fragment length matters more than a moderate error rate (\rev{Sect.~\ref{sec:sizecross} and} \rev{Sect.}~\ref{app:size}).

\subsection{T3: documentary versus literary}
\begin{figure}[!ht]
\centering
\includegraphics[width=\columnwidth]{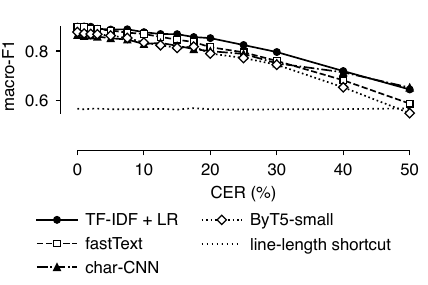}
\caption{T3 documentary versus literary: macro-F1 of every model against CER, with the line-length-only shortcut}
\label{fig:t3}
\end{figure}

Telling a document from a literary text is the most robust task in the study \rev{(Fig.~\ref{fig:t3})}. All four models score between 0.865 and 0.898 macro-F1 on clean text and lose little as the text degrades: TF-IDF keeps 0.852 at 20\% CER and 0.796 at 30\%, holding 90\% of its clean score up to 20\% CER, \rev{the highest tolerance in the study; document type, the next most tolerant task, reaches 12.5\% with its best model} (\rev{Sect.}~\ref{app:thresholds}). The signal is lexical and spread over the whole text, which is why scattered errors cost so little; the control model that sees only line-length statistics stays at 0.57 whatever the CER, so the number of lines and amount of letters do not give the answer away. The real limit is length rather than error rate: fragments under 20 letters are near chance even on clean text, while texts of 500 letters or more keep 90\% of their score up to \rev{17.5--25\%} CER (\rev{Sect.}~\ref{app:size}).

\subsection{T4: search}
\begin{figure}[!ht]
\centering
\includegraphics[width=\columnwidth]{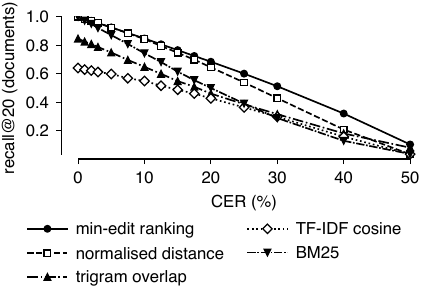}
\caption{T4 search: document-level recall@20 of the ranking methods over all 2,534 queries against CER}
\label{fig:t4}
\end{figure}

Search separates into ranking \rev{(Fig.~\ref{fig:t4})} and filtering. Ranking by minimum edit distance is the most robust and the most predictable method in the study: document-level recall@20 falls almost linearly, by about 1.6 points per point of CER, from \rev{0.92} at 5\% CER to 0.68 at 20\% and 0.51 at 30\%. Term-weighting schemes fare worse under noise than their clean scores suggest. BM25, tuned on clean validation queries, chose 6-letter $n$-grams and so behaves almost like exact containment: perfect on clean text, but 0.29 at 30\% CER against 0.51 for edit distance, a reminder that tuning a retrieval method on clean text selects for brittleness. Filtering fails differently: exact match keeps its precision as the text degrades (0.84 at 20\% CER) and loses recall instead, which is the safer failure for a scholar asking whether a formula occurs at all. Query length drives everything --- queries of \rev{10--12} letters tolerate 20\% CER, queries of three to six letters only 3\% (\rev{Sect.}~\ref{app:size}) --- and formulae survive far better than dictionary stems, so searching noisy text for a long phrase stays reliable long after single-word search has failed.

\section{Cross-task findings}\label{sec:cross}

\subsection{Lost lines}
Missing a line altogether is a different failure from misreading one, and it costs far less. Dropping 30\% of the text-bearing lines is worth about as much as \rev{5--10\% CER} for the document-level tasks, while for search it removes recall in proportion to the text lost. The ablation and its figure are in \rev{Sect.}~\ref{app:lost}.

\subsection{Fragment size}\label{sec:sizecross}
\rev{The length of a fragment limits every task more than a moderate error rate does (Fig.~\ref{fig:size}, Table~\ref{tab:sizecmp}). The reason is arithmetic. The models pool evidence over the letter $n$-grams of a document; the number of $n$-grams grows with its length, and scattered errors at a rate \cer{} leave about $(1-\cer)^n$ of them intact, which at 20\% CER is half of the trigrams and a third of the 5-grams. A text of 500 letters at 20\% CER therefore still offers some 250 intact trigrams, five times as many as an error-free fragment of 50 letters contains at all. Noise removes a fraction of the evidence; shortness removes most of it.

The tasks are evaluated on documents of different lengths (a median of 410 letters for T1, 134 for T2 and 122 for T3), but compared inside the same size bins they keep their order: from 50 letters upward dating has the lowest threshold in every bin (Table~\ref{tab:sizecmp}). Short fragments matter most, because unpublished material is more fragmentary than what editors have chosen to publish: 44\% of our documents with letters have fewer than 100 letters. On such fragments every task starts lower or tolerates less. The best dating model errs by 80 years on error-free text, against 55 overall, and the \cninety{} of the best classifier is 3\% under 100 letters against 15\% above for document type, and 10\% against 25\% for documentary versus literary. Sect.~\ref{app:size} gives the bins, the figure and the detail.}

\subsection{\revhead{The shape of the errors}}\label{sec:shape}
\rev{At a fixed CER the arrangement of the errors changes the outcome (the variants are defined in Sect.~\ref{sec:degrade}; Sect.~\ref{esm:shape} gives the curves and the thresholds). Bursts are the mildest shape: the same number of edited letters in runs of 2--4 touches fewer $n$-grams, and TF-IDF keeps 0.828 macro-F1 on document type at 20\% CER against 0.678 for scattered errors. Its \cninety{} moves from 12.5 to 17.5\% for document type and from 20 to 30\% for documentary versus literary. Substitutions alone are the harshest shape, and ranking by edit distance is the least sensitive to any of them. No variant lowers a threshold by more than one step of the CER grid, and bursts raise it by up to two. Among the shapes we tried the default scatter is thus close to the least favourable. Real recognition errors cluster and so resemble bursts more than an even scatter, which suggests that our thresholds err on the conservative side. Sect.~\ref{sec:real} confirms this on real output, where a quarter of the errors lie in runs longer than the bursts of this variant.}

\section{\revhead{Validation on real HTR output}}\label{sec:real}

\rev{The thresholds above rest on synthetic noise. This section sets that noise against the output of a real recogniser and asks one question: with the same number of errors, how do the four tasks behave on real HTR output compared with our synthetic noise?}

\subsection{\revhead{Design: synthetic twins for every papyrus}}
\rev{Each papyrus of the validation set gets two kinds of noisy version of the same text. One is the transcription of its images by Anagnostes~\cite{anagnostes2026}, a recogniser developed by the authors of this paper. The other is a set of five \emph{synthetic twins}: the clean L2 text of the same lines, degraded by the default noise of Sect.~\ref{sec:degrade} with exactly as many edits as separate the real output from the edition, under five seeds. The papyrus, its metadata, its length and its number of errors are the same in every version, so a difference in score can only come from real errors behaving differently from scattered ones.

The twins are needed because binning the real output by CER and laying it over the curves of Sect.~\ref{sec:results} would compare different documents. A bin of real output has its own mix of lengths, types and dates, and in real output the papyri with a high CER are the damaged, cursive and short ones, a link that synthetic noise does not have. Pairing removes both confounds and gives a much narrower interval than a bin of a few dozen papyri.}

\subsection{\revhead{Papyri and protocol}}\label{sec:realset}
\rev{\textbf{Papyri.} A papyrus enters only if Anagnostes has never met it in any training round, as an image or as text, and if its image is available online and the holding collection allows it to be used. Document type is cross-validated, so each papyrus is scored by the fold model that did not train on it; the other tasks use papyri of the test split only. Of 1,009 papyri attempted (Sect.~\ref{esm:protocol}), 459 have usable images; a papyrus can have several. We keep only those with a CER below 50\%, at least three matched lines (lines of the edition that Anagnostes segmented and read closely enough to be aligned with them) and at least 50 letters on them: 373 papyri (318 documentary, 55 literary) in 889 images from 18 collections, with 7,584 matched lines and 183,825 letters on them. The ground truth for each papyrus is its edition in papyri.info. We test document type on 220 of them, dating on 163, documentary versus literary and search on 225.

\textbf{Recognition and scoring.} Each image is read end to end by the line detector and the recogniser of the system, with no language model and no correction of the output. The output is reduced to L2 and aligned to the edition line by line, and the CER of a papyrus is the number of edits over the letters of its matched lines (Sect.~\ref{sec:cerdef}). It ranges from 1.2 to 49.7\%, with a median of 22.1\% (quartiles 14.9 and 31.0\%). Lines of the edition that Anagnostes did not segment, or read too badly to be aligned, are lost lines (a median of 11\% of the lines of a papyrus) and are left out of the twins. The frozen models of Sect.~\ref{sec:models} are applied to the clean text of the matched lines, to the real output and to the twins; search runs over the matched lines of the 225 test papyri with the 1,073 queries that occur in their clean text. Differences come with a 95\% interval from a paired bootstrap over papyri (queries for search). The line detector and the three conditions on the papyri were chosen after first results had been seen; Sect.~\ref{esm:protocol} gives the protocol in full and the results without these choices, which lead to the same conclusions.}

\subsection{\revhead{Real output against synthetic noise}}\label{sec:realresults}
\begin{figure*}[!t]
\centering
\includegraphics[width=\textwidth]{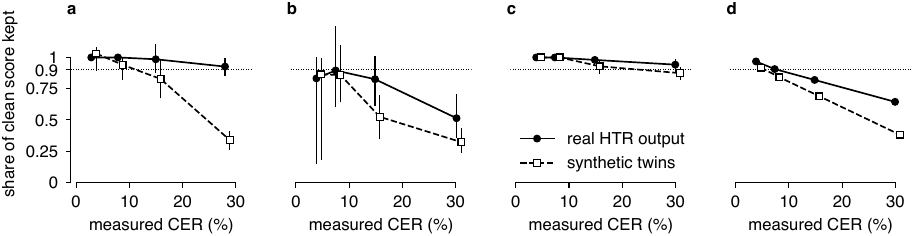}
\caption{\rev{Real HTR output and its synthetic twins by measured CER: the share of the score on clean text that is kept (retention, Sect.~\ref{sec:thresholds}), with 95\% intervals; the dotted line marks 0.9. The papyri are grouped by measured CER (0--5, 5--10, 10--20 and 20--50\%) and each group is drawn at its median CER. \textbf{a} T1 document type (TF-IDF, macro-F1). \textbf{b} T2 dating (fastText, MAE). \textbf{c} T3 documentary versus literary (TF-IDF, macro-F1). \textbf{d} T4 search (exact-match filter, F1 over lines). Intervals that leave a panel are cut at 1.25}}
\label{fig:real}
\end{figure*}

\rev{With the same number of errors, real output costs the tasks less than the synthetic noise (Table~\ref{tab:real}). For document type the TF-IDF model reaches a macro-F1 of 0.861 on the clean text, 0.823 on the real output and 0.501 on the twins (difference 0.32, 95\% interval 0.25--0.39). fastText and ByT5-small show the same gap; the char-CNN, the model least affected by synthetic noise, shows a smaller one whose interval reaches zero. The same is seen in dating: every model dates the real output better than the twins, and for the best model (fastText) real errors add 27 years to the error on clean text where the twins add 62. Documentary versus literary loses little in either version, and only TF-IDF separates them. In search, an exact match finds the lines of a query with an F1 of 0.72 in the real output and 0.53 in the twins, and every ranking method ranks better on the real output. Only the approximate filters, which accept a hit within one or two edits or within a normalised distance of 0.25, do slightly worse there (Sect.~\ref{esm:protocol}).

The gap opens with the CER (Fig.~\ref{fig:real}; Table~\ref{tab:realbycer}). Below 10\% CER real output and twins cannot be told apart in the three document tasks, but these groups hold only 7 to 22 papyri per task and would not show a difference of the size seen above. From 10\% on the two diverge: at 20--50\% CER (median 28\%) the document-type classifier still keeps 93\% of its clean score on real output (interval 85--99\%) and 34\% on the twins. Search has enough queries in every group and separates already below 5\% CER. The validation therefore supports the thresholds in their direction between 10 and 50\% CER for the document tasks and over the whole range for search; it cannot confirm the exact values of the thresholds that lie below 10\%.

Read as statements about real output, the thresholds of Sect.~\ref{sec:results} are thus lower bounds: at a given CER the real output of this recogniser supports each task at least as well as the synthetic noise does. The order of the tasks is broadly the same in both. With their best model, documentary versus literary and document type keep more than 90\% of the clean score in every CER group; exact search keeps 91\% at 5--10\% CER and 82\% at 10--20\%; dating keeps 83\% at 10--20\% and 51\% above, and remains the task that asks most of the recogniser.}

\begin{table}[!t]
\caption{\rev{Real HTR output against synthetic twins with the same number of edits. Clean: the score on the clean text of the matched lines. Last column: the score on real output minus the mean score on the five twins, with its paired 95\% bootstrap interval; for dating a negative difference means that real output is dated better. Search is scored on lines (trigram overlap and min-edit ranking as in Sect.~\ref{sec:models}); R@20: recall@20}}
\label{tab:real}
\centering\footnotesize\setlength{\tabcolsep}{2.5pt}\revcolor
\begin{tabular}{@{}lcccc@{}}
\toprule
 & Clean & Real & Twins & Real $-$ twins (CI) \\
\midrule
\multicolumn{5}{@{}l}{\textit{T1 document type, macro-F1, 220 papyri}} \\
TF-IDF & 0.861 & 0.823 & 0.501 & $+0.32$ ($+0.25$, $+0.39$) \\
fastText & 0.797 & 0.714 & 0.434 & $+0.28$ ($+0.20$, $+0.36$) \\
char-CNN & 0.739 & 0.649 & 0.580 & $+0.07$ ($0.00$, $+0.14$) \\
ByT5-small & 0.767 & 0.680 & 0.384 & $+0.30$ ($+0.22$, $+0.37$) \\
\addlinespace
\multicolumn{5}{@{}l}{\textit{T2 dating, MAE in years, 163 papyri}} \\
TF-IDF & 73 & 107 & 129 & $-22$ ($-35$, $-11$) \\
fastText & 39 & 66 & 101 & $-35$ ($-50$, $-19$) \\
char-CNN & 45 & 91 & 106 & $-15$ ($-26$, $-4$) \\
ByT5-small & 56 & 86 & 119 & $-34$ ($-50$, $-17$) \\
\addlinespace
\multicolumn{5}{@{}l}{\textit{T3 documentary versus literary, macro-F1, 225 papyri}} \\
TF-IDF & 0.975 & 0.936 & 0.880 & $+0.06$ ($+0.02$, $+0.09$) \\
fastText & 0.962 & 0.937 & 0.927 & $+0.01$ ($-0.02$, $+0.04$) \\
char-CNN & 0.930 & 0.861 & 0.867 & $-0.01$ ($-0.06$, $+0.03$) \\
ByT5-small & 0.929 & 0.885 & 0.851 & $+0.03$ ($0.00$, $+0.07$) \\
\addlinespace
\multicolumn{5}{@{}l}{\textit{T4 search, 1,073 queries over 4,444 lines}} \\
Exact match, F1 & 1.000 & 0.717 & 0.529 & $+0.19$ ($+0.18$, $+0.20$) \\
BM25, R@20 & 1.000 & 0.591 & 0.345 & $+0.25$ ($+0.22$, $+0.27$) \\
Trigram, R@20 & 0.997 & 0.665 & 0.503 & $+0.16$ ($+0.14$, $+0.18$) \\
Min-edit, R@20 & 1.000 & 0.689 & 0.608 & $+0.08$ ($+0.06$, $+0.10$) \\
\bottomrule
\end{tabular}
\end{table}

\subsection{\revhead{Why real errors cost less}}\label{sec:profile}
\rev{The versions of a papyrus carry the same number of errors, so the difference must lie in how the errors are arranged. In the real output three quarters of the errors touch another error and a quarter lie in runs of five or more; the default noise places every error on its own. We believe that this is what protects the tasks: runs spare the text between them. Of the four-letter sequences of the edition, 60\% survive in the real output of the same line and 37\% in the twins (49\% against 20\% above 20\% CER), and 14\% of the lines carry no error at all, against 6\%. A model that reads character $n$-grams, or a query that must match letter for letter, finds more of what it needs. Real output also drops more letters and gathers its errors beside lost text. Sect.~\ref{esm:protocol} gives the full profile (Table~\ref{tab:profile}) and sets it beside the burst variant of Sect.~\ref{sec:shape}, which comes closest to it without reaching it.}

\section{Discussion}\label{sec:discussion}

\subsection{What HTR quality is required for what}
Table~\ref{tab:targets} turns the thresholds into targets, read against what current systems report \rev{for papyri: an average near 18\%, with the best results at 2--9\% and 21--26\% on abraded lines and lines with lacunae~\cite{anagnostes2026announcement}}. Three groups emerge. Documentary-versus-literary classification works on today's output as it is; document type and search require \rev{accurate} HTR output \rev{if its errors are scattered as in our noise}. Dating needs either very \rev{accurate} HTR or a better model, since the text alone does not give a century-level answer even when \rev{transcribed without error. This suggests that dating on top of HTR should not rest on the text alone but also use the image: for Greek papyri a pipeline that dates fragments from the shapes of their letters has been proposed~\cite{west2024dating}, and for Latin inscriptions a multimodal model combines image and text for geographical attribution~\cite{assael2025aeneas}}. \rev{The validation of Sect.~\ref{sec:real} moves these targets in one direction only: on the real output of Anagnostes the same number of errors costs less, so the targets are lower bounds (last two columns of Table~\ref{tab:targets}). Document type, which the synthetic thresholds place beyond typical output for cursive hands, kept 93\% of its clean score at 20--50\% CER with a classifier trained on the editions, and exact search kept 82\% at 10--20\%. Dating remains the bottleneck on real output as well.}

\begin{table}[!ht]
\caption{CER targets per task, read from the per-task results of Sect.~\ref{sec:results} and the per-model thresholds of \rev{Sect.}~\ref{app:thresholds}. ``As is'' is the highest CER at which \rev{models} built on editions \rev{keep} 90\% of \rev{their performance on synthetic noise: the task-level \cninety{} of Sect.~\ref{sec:thresholds} and, in brackets, that of the best single model. ``Real output'' is the share of its clean score that the model of Fig.~\ref{fig:real} kept on the real output of Anagnostes, for papyri read with a CER of 10--20\% and of 20--50\% (Sect.~\ref{sec:realresults}; search over lines, ranking by minimum edit distance; subtypes were not validated). For real HTR output the thresholds are lower bounds}}
\label{tab:targets}
\centering\footnotesize\setlength{\tabcolsep}{2pt}
\begin{tabular}{@{}>{\raggedright\arraybackslash}p{1.95cm}ccc>{\raggedright\arraybackslash}p{1.6cm}@{}}
\toprule
 & & \multicolumn{2}{c}{\rev{Real output}} & \\
\cmidrule(lr){3-4}
Task & As is & \rev{10--20\%} & \rev{20--50\%} & Also depends on \\
\midrule
Doc.\ vs literary & 20\% \rev{(20\%)} & \rev{98\%} & \rev{94\%} & $\geq$50 letters \\
Document type & \rev{7.5\% (12.5\%)} & \rev{99\%} & \rev{93\%} & $\geq$100 letters \\
\rev{Document subtype} & \rev{5\% (10\%)} & \rev{--} & \rev{--} & \rev{as document type} \\
Search, ranking & 5\% \rev{(5\%)} & \rev{84\%} & \rev{62\%} & query length (20\% at \rev{10--12} letters) \\
Search, exact filter & 5\% & \rev{82\%} & \rev{64\%} & recall falls with CER, precision holds \\
Dating & 3\% \rev{(3\%)} & \rev{83\%} & \rev{51\%} & text length\rev{; no century-level answer} \\
\bottomrule
\end{tabular}
\end{table}

\subsection{Recommendations}
For HTR developers: report a letters-only CER computed after the normalisation of Sect.~\ref{sec:cerdef}, and report it by fragment size and the line-loss rate separately, because each of these changes what the CER means downstream. For tool builders: \rev{test a tool at a known CER with the released degradation algorithm and frozen corpora, and do not tune it on clean text alone: BM25, tuned on clean queries, chose long $n$-grams and lost more than edit-distance ranking on synthetic noise and on real output alike}. For papyrologists and collection curators: automatic first-pass transcription over unedited material at 10--20\% CER already supports \rev{a first sorting of a collection: documentary and literary fragments can be told apart, so that papyrologists find the pieces that belong to their own research agenda sooner (what to publish remains their choice),} and \rev{cataloguing by document type, for which a classifier trained on editions kept a macro-F1 of 0.82 on real output with a median CER of 21\%}; searching this transcription layer with edit-distance ranking and queries of ten letters or more is reliable; dating established from such transcription should be treated as a hint. \rev{For a first rough classification it is usually enough to assign a papyrus to the Ptolemaic, Roman or Byzantine period, which asks for a precision of about $\pm$150 years.}

\subsection{Limitations}
\rev{\textbf{The validation.} The thresholds are measured on simulated noise and compared with real output only within limits (Sect.~\ref{sec:real}): one recogniser, developed by the authors; 373 papyri whose images are available online and may be used; and measured CERs of 1--50\%, with only 44 papyri below 10\%. The exact values of the thresholds below 10\% CER, where those of dating, search and document type lie, could therefore be tested for search only. The CER of the validation is measured on lines that could be matched to the edition; a line read too badly to be matched counts as a lost line, and a user of the raw output meets both. The line detector and the conditions on the papyri were chosen after first results had been seen; the results without these choices agree (Sect.~\ref{esm:protocol}).}

\rev{\textbf{The noise.} Real errors are shaped by a combination of letter confusions, the hand, the damage and the recogniser's early stage of development. They come in runs and gather beside lost text, where the default noise spreads them evenly, over the text and over the documents: every document receives the CER of the corpus, whereas in real output the CER varies from one papyrus and one hand to another. The variants only bracket this, and they were not run for dating, the task with the lowest threshold. For another system, another kind of hand or a CER outside the validated range, the thresholds should be re-read on that system's own output, which the released pipeline makes a matter of replacing one step.} Only letter errors and lost lines are modelled, not merged or split lines, wrong reading order or fragments joined out of order. \rev{Three conventions of the letters-only view were held fixed and their effect was not measured: doubtful letters count as preserved, the gap token is given to the models, and search runs on text without word division.}

\rev{\textbf{Labels, texts and models.} The labels are scholarly judgements: the dates are the editors' intervals, the document types follow one typology that covers a small selection of papyri, and all classifiers are closed-set. The test population consists of published papyri, which are often longer and better preserved than the unpublished fragments an HTR pipeline is meant for (Sect.~\ref{sec:sizecross}).} The tasks use small, self-trained models; a larger pretrained model may be more or less robust. \rev{Every model is trained on clean editions. The first version of this study also retrained the models on synthetically degraded text: this helped on synthetic noise, but on real output no retrained model did better than its clean-trained counterpart, so the regime is not reported here. Training on recognised text is not covered.} Finally, the study covers Greek only; \rev{a Coptic replication uses the same pipeline and is} in progress.

\rev{\textbf{Further work.} Dating should combine the text with the image: palaeographic dating from the handwriting~\cite{degregorio2024helldate,west2024dating} draws on evidence that recognition errors do not touch. Catalogues of unpublished collections usually hold keywords assigned when a piece was conserved and digitised; given to the models alongside the recognised text they should help document typing and search in particular, and in our setting they can be imitated by the content keywords of HGV matched to strings of the DDbDP text. Lemmatised resources such as Trismegistos Words~\cite{tmwords} would turn string search into search by lexeme, an open-set classifier would add the ``unknown'' outcome that cataloguing needs, and post-correction of the recognised text~\cite{pavlopoulos2024challenging} could be tested as a way of absorbing errors. How the state of the writing (faded, abraded or lost letters) shapes the errors of a recogniser, and a noise model conditioned on it and on the runs of real errors (Sect.~\ref{sec:profile}), on which models could also be trained, belong to the evaluation of the recogniser itself and are left to a separate study.}

\section{Conclusion}\label{sec:conclusion}

We asked how \rev{accurate} HTR of Greek papyri must be for the tasks papyrologists would want to run on it, and answered per task with controlled experiments on 63,846 edited texts reduced to what \rev{an error-free} HTR would return and degraded to exact error rates. The tolerance spans close to an order of magnitude: from 20\% CER for telling documents from literature to 3\% for dating, with document type identification and search between. \rev{On the real output of a current recogniser, on 373 papyri that it had never seen, the same number of errors does less damage than our scattered noise, which we attribute to real errors coming in runs: the thresholds are lower bounds on what real output allows, tested between 10 and 50\% CER and, for search, over the whole range.} \rev{An error-free} transcription is not needed, and \rev{one overall CER number} is not enough: what matters is the CER for the task at hand, \rev{the way the errors are arranged (scattered or in runs) and} the length of the fragment. The corpus rules, the degradation algorithm, the frozen corpora, the models and every result table are released so that the thresholds can be re-read as HTR systems improve and \rev{on the output of other systems}.

\backmatter

\section*{Statements and Declarations}

\bmhead{\revhead{Supplementary information}}
\rev{\ESM{} (ESM\_1.pdf) holds the construction of the search queries and the list of formulae, the rendering rules of the letters-only text, the distribution of the dated documents over the centuries, the degradation algorithm in full, the models and their frozen settings, the thresholds of every model, the analyses of error shape, lost lines and fragment size with their tables, and the protocol and further results of the validation on real HTR output.}

\bmhead{Funding}
No funding was received for conducting this study.

\bmhead{Competing interests}
\rev{Anagnostes, the recogniser whose output is used for the validation of Sect.~\ref{sec:real}, was developed by the authors. Apart from this,} the authors have no competing interests to declare that are relevant to the content of this article.

\bmhead{Ethics approval and consent}
Not applicable; the study uses published editions and public metadata and involves no human participants.

\bmhead{Data availability}
The source data are public: the papyri.info idp.data repository (CC-BY, commit \texttt{fd88c0c} \rev{of 6 July 2026}~\cite{idpdata2026}), the Grammateus export of 13 September 2026 \rev{(CC BY-NC)}~\cite{grammateus2023}, and the OCR text of the Preisigke supplements on the Internet Archive~\cite{preisigke1971supplement}, which is in copyright and is fetched, not redistributed; every input is pinned by commit or SHA-256 in the released configuration. The derived corpora, result tables, bootstrap curves and thresholds are released with the code\rev{; the Grammateus labels are not redistributed, the pipeline joins them from the project's export}. \rev{The validation set of Sect.~\ref{sec:real} is released with it: the identifiers of the papyri, the Anagnostes transcriptions, their alignment to the lines of the editions and the scores; the synthetic twins are rebuilt by the released code. The images, the links to them and the text of the editions are not redistributed.}

\bmhead{Code availability}
Reference code for the experiments reported here---corpus construction from EpiDoc, the letters-only view, the exact-CER degradation, the four tasks and their models, the eight search methods and the bootstrap analysis---is at \url{https://github.com/repushko/good_enough_papyrus_htr}. The complete pipeline (rendering, normalisation, degradation, models, search, analysis and report) is available in addition, with a Makefile that reproduces every step from the pinned inputs. Task identifiers in \rev{the complete pipeline} differ from the numbering used here (documentary versus literary is T4 and search is T5 there)\rev{, and in both repositories the letters-only level L2 of Sect.~\ref{sec:cer} is called L3}.

\bmhead{\revhead{Acknowledgements}}
\rev{The validation of Sect.~\ref{sec:real} reads images of papyri that their holders make available online. We gratefully acknowledge the digitisation work undertaken by all the collections whose papyri were used in this validation. The images stay with these collections and are not redistributed.}

\bmhead{Author contributions}
Elena Chepel formulated the papyrological tasks and defined the corpus rules. Anton Repushko designed and implemented the experimental pipeline---the degradation algorithm, the models, the search methods and the statistical analysis---and ran all experiments. Both authors wrote the manuscript and approved the final version.

\bibliography{references}

\clearpage
\twocolumn[{\centering
{\LARGE\bfseries \rev{Online Resource 1}\par}\vspace{8pt}
{\large \rev{Which papyrus HTR is good enough? Character-error-rate tolerance of four papyrological tasks on Greek texts}\par}\vspace{6pt}
{\normalsize \rev{International Journal on Document Analysis and Recognition}\par}\vspace{4pt}
{\normalsize \rev{Anton Repushko (Independent Researcher, Berlin, Germany; anton@repushko.com) and Elena Chepel (University of Vienna, Vienna, Austria)}\par}
\vspace{20pt}}]
\setcounter{section}{0}\setcounter{figure}{0}\setcounter{table}{0}
\renewcommand{\thesection}{S\arabic{section}}
\renewcommand{\thefigure}{S\arabic{figure}}
\renewcommand{\thetable}{S\arabic{table}}
\renewcommand{\theHsection}{esm.\arabic{section}}
\renewcommand{\theHfigure}{esm.\arabic{figure}}
\renewcommand{\theHtable}{esm.\arabic{table}}
\movednote{For review this Online Resource is kept in the same file as the paper and will be split off as ESM\_1.pdf. Section numbers without S refer to the paper.}

\section{\revhead{Search queries}}\label{esm:queries}\label{app:formulae}
\movednote{How the query sets are drawn: detail moved from Sect.~\ref{sec:t4data}, with the counts of the selection added. The list of formulae was Appendix~F.}

\rev{All queries are strings of 3 to 63 letters, without accents, capitals or spaces, and each has at least one relevant test document (Sect.~\ref{sec:t4data}).

\textbf{Dictionary stems (1,000).} A headword candidate is the first word of a line of the OCR text of the Preisigke supplements that is followed by a full stop or a comma, consists mostly of Greek letters and has at least four of them (18,478 candidates). Its stem is the headword with its longest inflectional ending removed, provided that four letters remain (16,868 distinct stems). A stem is kept if it occurs in clean training text and in at least one test document (8,887 stems), and 1,000 are drawn at random, evenly over six strata of test-document frequency (1, 2--3, 4--7, 8--15, 16--63, and 64 or more documents).

\textbf{Names (500).} The editions carry no name markup, so a name is a word form of at least four letters that editors capitalise in at least two training documents, not counting the first word of an edition or of a sentence (19,864 of 60,337 capitalised forms). 11,881 of them occur in a test document, and 500 are drawn at random over the same strata. They are proper nouns in general: people, places, gods and months.

\textbf{Random substrings (1,000).} For each length from 3 to 12 letters, 100 distinct substrings of clean test lines, never spanning a gap.}

\movednote{The next paragraph was moved from Sect.~\ref{sec:t4data}.}

\rev{\textbf{Alternatives not used.} The lemmatised word lists of Trismegistos Words~\cite{tmwords} or the PapyGreek treebanks~\cite{vierros2021papygreek} would be alternative sources to the Preisigke stems and would allow relevance by lemma instead of by string, which we leave for future work. Because queries and texts carry no word spaces, a query can also match across a word boundary, which a search on word-divided text would exclude; we did not search word-divided text, because the division is not on the papyrus and would itself have to be restored from the noisy transcription.}

\rev{\textbf{Formulae (34 of 35 used).} Thirty-five formulae were chosen by the authors as standard phrases of Greek documentary papyri: the greetings, health wishes and closing words of letters, dating clauses, and the acknowledgement, request and subscription clauses of contracts, receipts and petitions. They are printed as the authors wrote them, each with the number of test documents (of 5,961) whose clean text contains it; the queries themselves are the same phrases reduced to letters (Sect.~\ref{sec:t4data}). One formula does not occur in the test split and was not used.}

{\revcolor\raggedright
\gk{qa'irein} (512);
\gk{poll`a qa'irein} (5);
\gk{>err\~wsja'i se e>'uqomai} (33);
\gk{>'errwso} (152);
\gk{>err\~wsjai} (120);
\gk{>'etouc} (499);
\gk{<omologe\~i} (48);
\gk{<omologo\~umen} (34);
\gk{>ap'eqw par`a so\~u} (8);
\gk{pr`o m`en p'antwn} (17);
\gk{e>'uqoma'i se <ugia'inein} (2);
\gk{t`o prosk'unhm'a sou poi\~w} (8);
\gk{par`a t\~w| kur'iw| Sar'apidi} (2);
\gk{di`o >axi\~w} (16);
\gk{<'in'' >\char126 w bebohjhm'enoc} (1);
\gk{dieut'uqei} (26);
\gk{e>ut'uqei} (90);
\gk{>epid'edwka} (52);
\gk{seshme'iwmai} (6);
\gk{A>utokr'atoroc Ka'isaroc} (133);
\gk{>'egraya <up`er a>uto\~u} (30);
\gk{m`h e>id'otoc gr'ammata} (7);
\gk{<wc >et\~wn} (54);
\gk{>asp'azomai} (20);
\gk{>asp'azeta'i se} (11);
\gk{>ap'odoc} (58);
\gk{kaj`wc pr'okeitai} (14);
\gk{kur'ia <h suggraf'h} (0, not used);
\gk{>eperwthje`ic <wmol'oghsa} (8);
\gk{<upate'iac} (50);
\gk{>indikt'ionoc} (45);
\gk{t\~h| kur'ia| mou} (5);
\gk{k'uri'e mou} (26);
\gk{A>ur'hlioc} (185);
\gk{basile'uontoc} (6).
\par}

\section{\revhead{Dating: documents per century}}\label{esm:centuries}
\movednote{Table moved from Sect.~\ref{sec:tasks} (T2).}

\rev{Table~\ref{tab:centuries} gives the distribution over the centuries of the documents on which dating is evaluated (Sect.~\ref{sec:tasks}). Only three of them lie outside the span from the third century BCE to the eighth century CE (nine in all splits). 43\% are dated to an exact year, which bounds the share of predictions that can fall inside the editor's interval.}

\begin{table}[!ht]
\caption{\rev{T2: documents with a date interval of at most 100 years, by the century of the interval midpoint (midpoints binned in steps of 100 years). All splits, and the test split; counts include documents without letters (38 in the test split), which are not evaluated}}
\label{tab:centuries}
\centering\footnotesize\setlength{\tabcolsep}{5pt}\revcolor
\begin{tabular}{@{}lrrr@{}}
\toprule
Century & All & \% & Test \\
\midrule
IV BCE & 5 & 0.0 & 1 \\
III BCE & 4,029 & 8.7 & 413 \\
II BCE & 3,086 & 6.6 & 321 \\
I BCE & 1,477 & 3.2 & 128 \\
I CE & 4,561 & 9.8 & 490 \\
II CE & 13,400 & 28.8 & 1,357 \\
III CE & 6,997 & 15.1 & 708 \\
IV CE & 4,889 & 10.5 & 506 \\
V CE & 1,274 & 2.7 & 121 \\
VI CE & 3,002 & 6.5 & 310 \\
VII CE & 3,054 & 6.6 & 301 \\
VIII CE & 704 & 1.5 & 61 \\
IX CE & 4 & 0.0 & 2 \\
\midrule
Total & 46,482 & 100 & 4,719 \\
\bottomrule
\end{tabular}
\end{table}

\section{\revhead{What counts as ink}}\label{app:ink}
\movednote{Was Appendix~A. The four-line version of Fig.~\ref{fig:pipeline} is the figure of the first submission.}

Table~\ref{tab:ink} lists how each EpiDoc element of the edition is rendered in the \rev{preserved text (L1)} of Sect.~\ref{sec:cer}.

\movednote{The next paragraph was moved from Sect.~\ref{sec:cer}.}

\rev{In the letters-only level (L2), lower case and a single sigma are choices of encoding, not a loss of information: papyri are written in majuscule and cursive hands without a distinction of case, which the editions transcribe in lower case; the capitals of an edition are an editorial convention for proper names; and the medial, final and lunate sigma of printed Greek are typographic variants of one written letter, so folding them discards nothing about the script, which a transcription does not record in any case.}

\begin{table}[!ht]
\caption{What counts as ink: how EpiDoc elements of the edition are rendered in the \rev{preserved text (L1)}. Counts are element occurrences over the 72,623 Greek editions of DDbDP and DCLP}
\label{tab:ink}
\centering\footnotesize\setlength{\tabcolsep}{3pt}
\begin{tabular}{@{}p{3.1cm}rp{2.3cm}@{}}
\toprule
EpiDoc element & Count & In the \rev{preserved text} \\
\midrule
\texttt{supplied} \texttt{@reason=lost} & 721,518 & removed, gap token \\
\texttt{supplied} \texttt{@reason=omitted} & 17,257 & removed, no gap (never written) \\
\texttt{gap} & 797,539 & gap token \\
\texttt{choice}: \texttt{orig}/\texttt{sic} vs \texttt{reg}/\texttt{corr} & 143,904 & \texttt{orig}/\texttt{sic} kept (what the scribe wrote) \\
\texttt{expan} with \texttt{ex} & 903,536 & written letters kept, expansion dropped \\
\texttt{abbr} & 37,113 & kept \\
\texttt{unclear} & 763,272 & kept (ink present, reading doubtful) \\
\texttt{del}, \texttt{add}, \texttt{surplus}, \texttt{subst} & 70,011 & kept (ink) \\
\texttt{num} & 583,726 & letters kept (numerals are letters) \\
\texttt{app}: \texttt{lem} vs \texttt{rdg} & 47,457 & \texttt{lem} kept, \texttt{rdg} dropped \\
\texttt{g}, \texttt{am}, \texttt{note}, \texttt{figure}, \texttt{certainty} & 78,575 & removed (symbols, notes) \\
\texttt{handShift}, \texttt{milestone} & 31,423 & removed \\
\texttt{lb}, \texttt{l} & 1,070,210 & line break; \texttt{break="no"} joins the word \\
\texttt{space} & 15,661 & word boundary, no letters \\
\bottomrule
\end{tabular}
\end{table}

\rev{Figure~\ref{fig:pipelinefull} is the four-line version of Fig.~\ref{fig:pipeline}: lines 1--2 of the same letter, which are complete and need no restoration, are shown with lines 19--20.}

\begin{figure*}[!t]
\centering
\input{figures/fig1_full.tex}
\caption{A real papyrus from the edition to letters-only text and its degradation. Lines 1--2 and 19--20 of the letter of Apion to his father (BGU II 423 = Chrest.Wilck.\ 480) as encoded in the DDbDP edition (L0; square brackets enclose restored letters, dots mark uncertain ones), with the editorial layer removed (\rev{L1, preserved text}; \gap{} marks lost text), and reduced to letters (\rev{L2}), the text \rev{an error-free} HTR system would return. The last three rows are the same lines degraded by the study's algorithm to 5, 15 and 30\% CER \rev{(the number of edits is whole)} with the default 3:1:1 mix: a substituted letter is bold, an inserted letter bold and underlined, and a deleted letter is marked by a bold dot. Every edit costs one Levenshtein step and gap tokens are never touched}
\label{fig:pipelinefull}
\end{figure*}

\section{\revhead{The degradation algorithm}}\label{esm:degrade}
\movednote{The four rules, the edit log and the paragraph on short documents are moved from Sect.~\ref{sec:degrade}, where a summary remains.}

The \rev{L2} text is degraded under four constraints that make the CER exact and the errors well defined:

\begin{enumerate}
\item \textbf{Edit budget.} For a split (train, validation or test) and a target CER \cer{}, the budget is $K = \mathrm{round}(\cer \cdot N)$ edits, $N$ being the letters on the kept lines of the split, apportioned to documents in proportion to their letters \rev{by largest remainders}, so the split CER equals \cer{} exactly and each document \rev{of $n$ letters receives its share $\cer \cdot n$ rounded to a neighbouring whole number of edits}.
\item \textbf{Positions.} A document's edits land on a uniformly random, pairwise non-adjacent subset of its letters. Gap tokens are never edited, and letters on the two sides of a gap do not count as adjacent.
\item \textbf{Edit types.} Each edit is a substitution, an insertion or a deletion with probabilities 3:1:1. A substitution \rev{replaces the letter by a uniformly random different letter. An insertion keeps the letter and writes a new one immediately before it, drawn uniformly from the letters that differ from both of its new neighbours (the preceding letter and the edited letter itself). A deletion removes the letter; a letter equal to one of its neighbours is never deleted, and such an edit exchanges its type with another edit of the document. Both restrictions exclude doubled letters, the commonest way in which an insertion or a deletion could be undone, or explained, by a neighbouring edit.}
\item \textbf{Invariant.} \rev{The Levenshtein distance between the clean and the degraded letters of a document must equal its edit count, so that every edit costs exactly one step and none is cancelled by another. The rules above make cancellation rare but cannot exclude it, so the distance is computed for every document of every corpus and a line on which it falls short is redrawn with the same budget (at 5\% CER this happens in 0.4\% of the test documents, at 20\% in 6\%). After redrawing, no document in any of the 576 corpora violates the invariant.}
\end{enumerate}

\rev{\textbf{Edit log and seeds.}} Every edit is logged with its line, offset and type, so any truth defined on the clean text can be projected onto the degraded text\rev{. Every} random draw is seeded from the cell's parameters and the document identifier, so each document is reproducible independently of processing order.

\rev{\textbf{Short documents and quantisation.}\label{sec:quant} Edits are whole, so a document of $n$ letters can only have a CER that is a multiple of $1/n$: at a nominal 5\% CER a fragment of 12 letters receives one edit (8.3\%) and a fragment of 8 letters none. The corpus CER stays exact, but short fragments are either untouched or noisier than nominal. Table~\ref{tab:effcer} gives the CER that each size bin of the test split actually receives. From 100 letters upward it equals the nominal CER to within 0.1 points at every level. The shortest bins are cleaner than nominal at the low end of the grid: no document under 20 letters is edited at 1\% or 2\% CER, 80\% of them are untouched at 3\% and 42\% at 5\%, and 94\% of the documents of 20--49 letters are untouched at 1\%. Thresholds at or below 3\% CER, where dating sits, are therefore carried by documents of about 50 letters or more, and the size-bin curves of the shortest fragments (\rev{Sect.}~\ref{app:size}) should be read against the effective CER, not the nominal one.}

\section{\revhead{Models and frozen settings}}\label{app:settings}
\movednote{The full version of Table~\ref{tab:models}, moved from Sect.~\ref{sec:models}, merged with the settings of the former Appendix~E.}

\rev{This section gives the models of Table~\ref{tab:models} in full.} All hyperparameters are tuned once on clean validation text (for document type, on a fold held out from training; Sect.~\ref{sec:splits}) and frozen. Every frozen hyperparameter is listed in the \texttt{FROZEN} table of \texttt{models.py}, and every seed, CER grid, split fraction and bootstrap setting in \texttt{config.py}, of the released code.

\begin{description}
\item[TF-IDF + logistic / ridge (T1--T3).] Letter 1--5-gram TF-IDF with sublinear term frequency and $\ell_2$-normalised rows~\cite{pedregosa2011sklearn}; multinomial logistic regression with \rev{the L-BFGS solver} for the classes and ridge regression for the dates. $\min\mathrm{df}$, $C$ and the class weights are tuned per task (the chosen $C = 100$ sits on the grid edge), as is the ridge $\alpha$ for dating.
\item[fastText (T1--T3).] Supervised fastText~\cite{joulin2017fasttext} with one token per letter, so its word $n$-grams are letter $n$-grams (3--5); softmax loss, minimum count 1, 2M buckets, dimension 100--200, 30--100 epochs, learning rate 0.5--1.0 (grid widened once after the first tuning hit its edge, then frozen). Dating is cast as 25-year classes and the prediction is the probability-weighted bin midpoint; T3 is trained with the literary class oversampled to balance.
\item[Char-CNN (T1--T3).] 2.6M parameters~\cite{zhang2015charcnn}: letter embeddings (64), a $k{=}7$ convolution, three stages of two residual blocks with dilated $k{=}5$ convolutions (256 channels), masked max + mean pooling and an MLP head; inputs up to 4,096 letters. Learning rate $3\times10^{-3}$ (T1) or $10^{-3}$, square-root inverse-frequency class weights for T1 and T3, at most 20 epochs (T1: 80), early stopping on the validation split (T1: on a fourth fold) with patience 4 (T1: 8).
\item[ByT5-small (T1--T3).] The 218M-parameter byte-level encoder of ByT5-small~\cite{xue2022byt5} with masked mean pooling and a linear head; inputs up to 2,048 bytes ($\approx$1,000 Greek letters, above the 90th percentile of document length). AdamW~\cite{loshchilov2019adamw} with learning rate $10^{-4}$ (head $10^{-3}$), weight decay 0.01, 6\% warm-up, \rev{bfloat16 mixed precision}, gradient clipping at 1.0, length-bucketed batches of at most 48k tokens and early stopping on the validation split with patience 6 evaluations; 3 epochs, 8 for T2.
\item[Line-length shortcut (T3).] Gradient-boosted trees on line-length statistics only: does layout give the answer away?
\item[Search methods (T4).] Exact match; Levenshtein filters with at most 1 or 2 edits; normalised distance $d/m \leq 0.25$\rev{, whose hits are ranked by $d/m$}; ranking by minimum edit distance $d$ (semi-global, Myers' bit-parallel algorithm~\cite{myers1999bitvector}); trigram overlap; TF-IDF cosine over letter 2--4-grams; Okapi BM25~\cite{robertson2009bm25} over letter $n$-grams ($n = 6$, $k_1 = 0.3$, $b = 0$, all on grid edges after one widening, tuned on clean validation queries, IDF from clean training units). Matches never span a gap or a line break.
\end{description}

\section{\revhead{Thresholds per model}}\label{app:thresholds}
\movednote{Was Appendix~B.}

Table~\ref{tab:permodel} gives \cninetyfive{} and \cninety{} for every model on the headline metric, with the clean score and its 95\% interval. The full table with size bins, variants and lost lines is in the released results. \rev{The task-level rows are the model-averaged thresholds of Sect.~\ref{sec:thresholds}, the ones quoted in the abstract.}

\movednote{The next paragraph was moved from Sect.~\ref{sec:results} (T1).}

\rev{\textbf{Subtypes.} The 16-subtype variant is harder and less tolerant. On clean text TF-IDF reaches macro-F1 0.845, the char-CNN 0.703 and fastText 0.680, against a baseline of 0.028 (ByT5-small was not run on it). TF-IDF keeps 90\% of its clean score up to 10\% CER, then falls faster than on the four types, to 0.407 at 20\% and 0.050 at 30\%; the task-level threshold over the three models is 5\%, the figure quoted in the abstract.}

\begin{table*}[!t]
\caption{Per-model thresholds (\% CER) on the headline metric. Clean: point estimate and 95\% bootstrap interval at 0\% CER\rev{. Task-level rows average the retention of the models named}}
\label{tab:permodel}
\centering\footnotesize\setlength{\tabcolsep}{2.5pt}
\begin{tabular}{@{}lllcc@{}}
\toprule
Task & Model & Clean score & \cninetyfive{} & \cninety{} \\
\midrule
T1 & TF-IDF + LR & 0.909 (0.895--0.923) & 10 & 12.5 \\
T1 & fastText & 0.857 (0.839--0.873) & 5 & 7.5 \\
T1 & char-CNN & 0.811 (0.793--0.829) & 5 & 10 \\
T1 & ByT5-small & 0.825 (0.808--0.843) & 3 & 5 \\
\rev{T1} & \rev{task level, four models} & \rev{0.851 (0.838--0.863)} & \rev{5} & \rev{7.5} \\
\rev{T1 subtype} & \rev{TF-IDF + LR} & \rev{0.845 (0.826--0.862)} & \rev{5} & \rev{10} \\
\rev{T1 subtype} & \rev{char-CNN} & \rev{0.703 (0.678--0.728)} & \rev{3} & \rev{7.5} \\
\rev{T1 subtype} & \rev{fastText} & \rev{0.680 (0.651--0.704)} & \rev{2} & \rev{5} \\
\rev{T1 subtype} & \rev{task level, three models} & \rev{0.743 (0.723--0.758)} & \rev{3} & \rev{5} \\
T2 & fastText & 55 y (52--58) & 1 & 3 \\
T2 & char-CNN & 62 y (60--64) & 2 & 3 \\
T2 & ByT5-small & 72 y (70--75) & 2 & 3 \\
T2 & TF-IDF + ridge & 83 y (81--86) & 2 & 3 \\
\rev{T2} & \rev{task level, four models} & \rev{68 y (66--71)} & \rev{2} & \rev{3} \\
T3 & TF-IDF + LR & 0.898 (0.877--0.919) & 12.5 & 20 \\
T3 & fastText & 0.897 (0.874--0.918) & 10 & 17.5 \\
T3 & ByT5-small & 0.876 (0.853--0.900) & 7.5 & 17.5 \\
T3 & char-CNN & 0.865 (0.840--0.889) & 7.5 & 17.5 \\
\rev{T3} & \rev{task level, four models} & \rev{0.884 (0.863--0.903)} & \rev{10} & \rev{20} \\
T3 & line-length shortcut & 0.567 (0.538--0.600) & $\geq$50 & $\geq$50 \\
T4 & min-edit ranking, recall@20 & 1.000 & 3 & 5 \\
T4 & normalised distance, recall@20 & 1.000 & 3 & 5 \\
T4 & BM25, recall@20 & 0.999 & 1 & 3 \\
T4 & trigram overlap, recall@20 & 0.846 (0.836--0.856) & 1 & 3 \\
T4 & TF-IDF cosine, recall@20 & 0.639 (0.625--0.653) & 2 & 5 \\
\rev{T4} & \rev{task level, five ranking methods, recall@20} & \rev{0.897 (0.893--0.901)} & \rev{2} & \rev{5} \\
T4 & exact match, filter F1 & 1.000 & 2 & 5 \\
T4 & BM25, filter F1 & 0.849 (0.825--0.869) & 3 & 7.5 \\
\bottomrule
\end{tabular}
\end{table*}

\section{\revhead{The shape of the errors}}\label{esm:shape}
\movednote{Table, figure and detail moved from Sect.~\ref{sec:shape}, where a summary remains; the figure was the supplementary figure of the first submission.}

\begin{figure*}[!t]
\centering
\includegraphics[width=\textwidth]{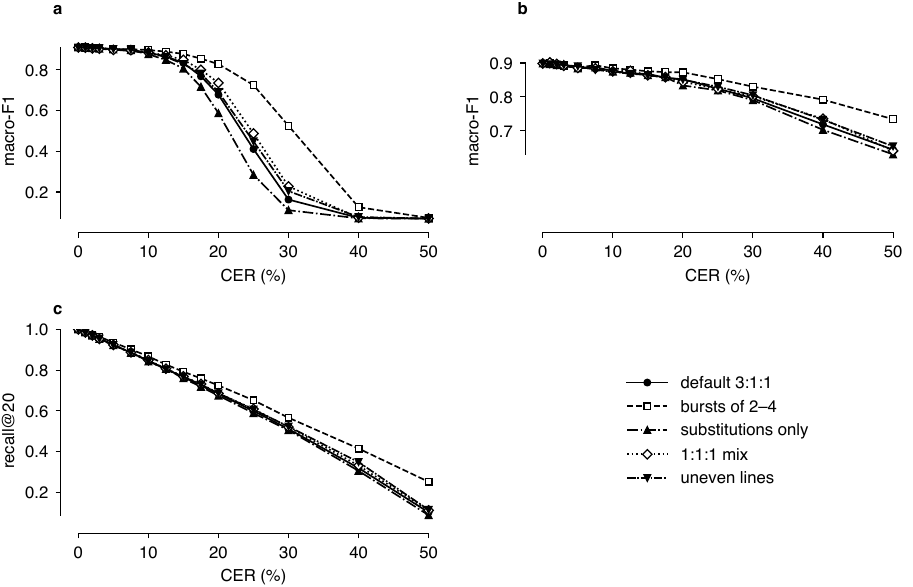}
\caption{\rev{The same edit budget spent in five ways. \textbf{a} T1 document type (TF-IDF + LR, macro-F1). \textbf{b} T3 documentary versus literary (TF-IDF + LR, macro-F1). \textbf{c} T4 search (min-edit ranking, document-level recall@20). The legend applies to all three panels}}
\label{fig:shape}
\end{figure*}

\rev{Table~\ref{tab:shape} and Fig.~\ref{fig:shape} show the same edit budget spent in the five ways defined in Sect.~\ref{sec:degrade}. Bursts are the mildest shape: the same number of edited letters in runs of 2--4 touches fewer $n$-grams and fewer words, and TF-IDF keeps 0.828 macro-F1 on document type at 20\% CER against 0.678 for scattered errors, and 0.525 against 0.163 at 30\%. Its \cninety{} moves from 12.5 to 17.5\% for document type and from 20 to 30\% for documentary versus literary. Substitutions alone are the harshest shape (0.587 on document type at 20\% CER) and the uniform mix is milder than the default (0.735), because an insertion leaves every original letter in place and destroys fewer $n$-grams than a substitution. Uneven lines stay close to the default (0.691). Ranking by edit distance is the least sensitive: its recall@20 at 20\% CER lies between 0.675 and 0.725 for all five variants.}

\begin{table}[!ht]
\caption{\rev{\cninety{} (\% CER) under the five error shapes of Sect.~\ref{sec:degrade}, all with the same number of edits. Search: recall@20 of min-edit ranking and F1 of the exact-match filter, document level}}
\label{tab:shape}
\centering\footnotesize\setlength{\tabcolsep}{2.5pt}\revcolor
\begin{tabular}{@{}lccccc@{}}
\toprule
 & Default & Subst. & Uniform & Bursts & Uneven \\
Task, model & 3:1:1 & only & 1:1:1 & of 2--4 & lines \\
\midrule
T1 type, TF-IDF & 12.5 & 12.5 & 15 & 17.5 & 12.5 \\
T1 type, fastText & 7.5 & 7.5 & 7.5 & 10 & 7.5 \\
T1 subt., TF-IDF & 10 & 10 & 10 & 15 & 10 \\
T1 subt., fastText & 5 & 3 & 5 & 5 & 3 \\
T3, TF-IDF & 20 & 20 & 20 & 30 & 25 \\
T3, fastText & 17.5 & 17.5 & 20 & 25 & 17.5 \\
T4, min-edit rank & 5 & 5 & 5 & 5 & 5 \\
T4, exact filter & 5 & 5 & 5 & 7.5 & 5 \\
\bottomrule
\end{tabular}
\end{table}

\section{\revhead{Lost lines}}\label{app:lost}
\movednote{Was Appendix~C.}

\begin{figure*}[!t]
\centering
\includegraphics[width=\textwidth]{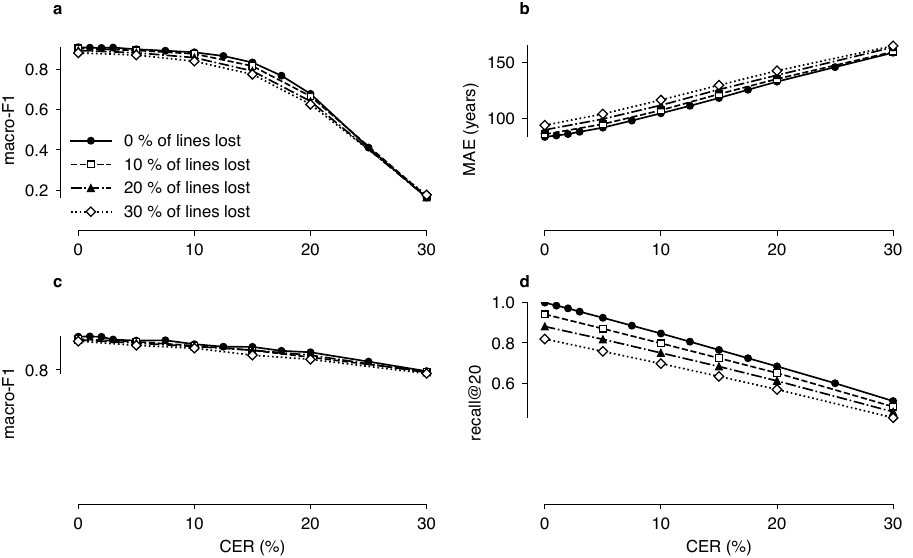}
\caption{Lost lines crossed with CER: 0, 10, 20 and 30\% of the text-bearing lines dropped. \textbf{a} T1 document type (TF-IDF + LR, macro-F1). \textbf{b} T2 dating (TF-IDF + ridge, MAE in years, lower is better). \textbf{c} T3 documentary versus literary (TF-IDF + LR, macro-F1). \textbf{d} T4 search (min-edit ranking, document-level recall@20). All four panels share the legend}
\label{fig:lost}
\end{figure*}

A recogniser can fail in two ways that a single CER figure does not distinguish: it can read a line and get its letters wrong, or it can miss the line altogether, because the layout analysis did not find it or because the ink is too faint to segment. The second failure removes text rather than corrupting it, so it is modelled separately (Sect.~\ref{sec:degrade}): a seeded fraction $p$ of a document's text-bearing lines is dropped before the letter noise is applied, and the CER is measured on the lines that remain. The two losses therefore compose without being confounded. Figure~\ref{fig:lost} crosses $p \in \{0, 10, 20, 30\}\%$ with $\cer \in \{0, \dots, 30\}\%$ for each task, using \rev{the TF-IDF model of each document-level task and min-edit ranking for search}.

For the three document-level tasks, losing lines costs strikingly little. On clean text, dropping 30\% of the lines lowers document type from 0.909 to 0.881 macro-F1 and documentary versus literary from 0.898 to 0.885, and raises the dating error from 83 to 94 years --- \rev{about what 10\% CER costs on its own for document type and 5--7.5\% CER for the other two}, for a third of the text removed. The curves in panels \textbf{a}--\textbf{c} stay close to parallel as the error rate rises, so the two kinds of loss add up almost independently rather than compounding; above 20\% CER they converge, because a text that \rev{is so} noisy carries little signal whether or not a third of it is missing.

Search behaves differently (panel \textbf{d}). A lost line cannot be found, so recall falls in proportion to the text removed rather than with the difficulty of matching: min-edit recall@20 drops from 1.000 to 0.820 on clean text when 30\% of the lines are gone, the largest single effect of line loss in the study, and the gap stays roughly constant at every CER.

The practical consequence is that a layout stage that misses a third of the lines is a much smaller problem for classification and dating than a recogniser that gets a fifth of the letters wrong, while for search the two matter about equally, since one removes what the other corrupts. This is the argument for reporting the line-loss rate alongside the CER (Sect.~\ref{sec:discussion}).

\section{\revhead{Fragment size}}\label{app:size}
\movednote{Was Appendix~D.}

\begin{figure*}[!t]
\centering
\includegraphics[width=\textwidth]{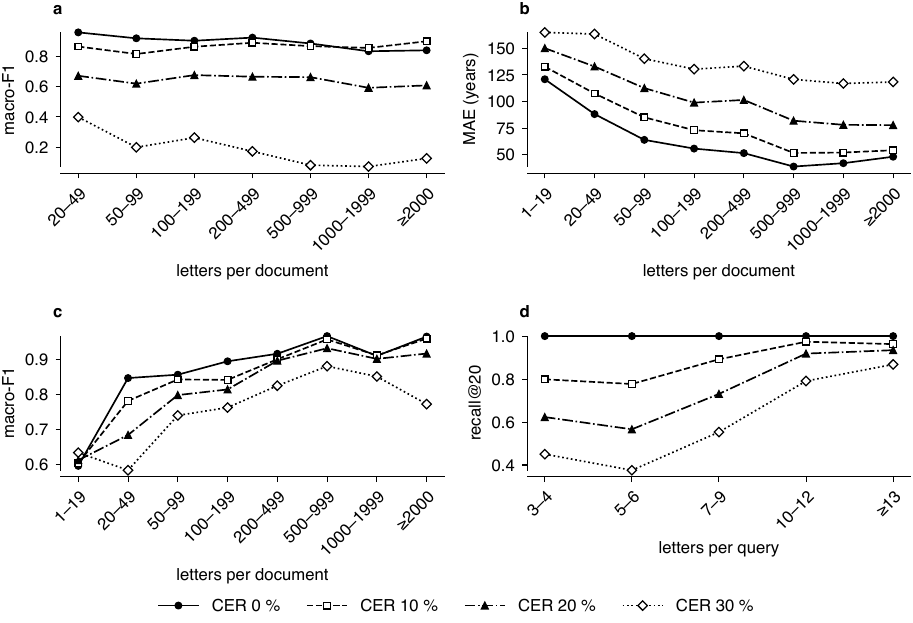}
\caption{Performance by fragment size at 0, 10, 20 and 30\% CER. Bin sizes are in the released result tables; two of the plotted bins hold fewer than 30 units and get no threshold (document type at 20--49 letters, 17 papyri; queries of 13 letters or more, 15 queries). \textbf{a} T1 document \rev{type} (TF-IDF + LR) by letters per document. \textbf{b} T2 dating error (char-CNN) by letters per document. \textbf{c} T3 documentary versus literary (TF-IDF + LR) by letters per document. \textbf{d} T4 recall@20 (min-edit ranking) by letters per query}
\label{fig:size}
\end{figure*}

Every task has a size below which no CER helps and a size above which a large CER is survivable (Fig.~\ref{fig:size}). Fragments under 20 letters are dated with \rev{an error of 121 to 141 years, depending on the model,} and are at chance for documentary versus literary on clean text; \rev{with TF-IDF, texts of 500 letters or more keep 90\% of their score up to 10--15\% CER for document type and up to 17.5--25\% for documentary versus literary (Table~\ref{tab:sizecmp}); on documents of 2,000 letters or more the four models range from 10 to 40\% for the latter.} Search follows the \rev{length of the} query rather than \rev{that of} the document: 10--12-letter queries tolerate 20\% CER, 3--6-letter queries 3\%. One counter-intuitive detail: on document type the longest documents collapse hardest \rev{with TF-IDF} at 30\% CER (0.071 macro-F1 for 1,000--1,999 letters, Fig.~\ref{fig:size}a), because they accumulate the largest number of unseen noisy $n$-grams.

\movednote{The next two paragraphs were moved from Sect.~\ref{sec:sizecross}.}

\rev{The tasks are evaluated on documents of different lengths (a median of 410 letters for T1, 134 for T2 and 122 for T3), so Table~\ref{tab:sizecmp} compares them inside the same size bins. From 50 letters upward dating has the lowest threshold in every bin (2--3\%). Document type and documentary versus literary reach 10--15\% and 7.5--25\% from 100 letters upward, and the latter is the more tolerant of the two in every shared bin but one. The order of the tasks is therefore not an artefact of their length distributions.

Short fragments deserve separate mention, because unpublished material is more fragmentary than what editors have chosen to publish. Of the documents with letters in our corpus, 9\% have fewer than 20 letters, 23\% fewer than 50 and 44\% fewer than 100. On error-free text the best dating model errs by 141 years on fragments under 20 letters and by 87 years on those of 20--49 letters, against 55 years overall, and documentary versus literary is near chance under 20 letters (macro-F1 0.60). If the test set is reduced to fragments of fewer than 100 letters (40\% of the dated test documents), the dating error of the best model rises from 55 to 80 years on error-free text and from 70 to 101 years at 10\% CER, while its threshold stays at 1--3\% in each of these bins (Table~\ref{tab:sizecmp}). The two classification tasks lose tolerance in the same way. We refitted their best model (TF-IDF) with the frozen settings to obtain predictions per document; the refit reproduces the reported scores to within 0.006. The 128 typed papyri under 100 letters are classified as well as the longer ones on error-free text (macro-F1 0.92 against 0.91), but at 10\% CER they keep 89\% of that score where the longer ones keep 98\%, and their \cninety{} is 3\% against 15\% (fastText: 3\% against 7.5\%). For documentary versus literary, the 2,575 test documents under 100 letters (43\%) start lower (0.83 against 0.92) and their \cninety{} is 10\% against 25\% (fastText: 10\% against 20\%); under 50 letters it is 2\% (fastText: 7.5\%).}

\rev{Table~\ref{tab:sizecmp} puts the three document-level tasks side by side in the same size bins, each with its best model (Sect.~\ref{sec:sizecross}), and Table~\ref{tab:effcer} gives the CER that the documents of each size actually receive at a nominal CER (Sect.~\ref{sec:degrade}). The two belong together: below 5\% nominal CER the thresholds of the bins under 50 letters are read on text that is cleaner than nominal.}

\begin{table*}[!t]
\caption{\rev{The document-level tasks in the same size bins: evaluated documents $n$, score on clean text with its 95\% bootstrap interval, and \cninety{} (\% CER) of the best model of each task. Bins with fewer than 30 documents get no threshold. The bracketed threshold belongs to a bin that is near chance on clean text and carries no information}}
\label{tab:sizecmp}
\centering\footnotesize\setlength{\tabcolsep}{4pt}\revcolor
\begin{tabular}{@{}llrrrrrrrr@{}}
\toprule
 & & \multicolumn{8}{c}{Letters per document} \\
\cmidrule(l){3-10}
Task (model) & & 1--19 & 20--49 & 50--99 & 100--199 & 200--499 & 500--999 & 1,000--1,999 & $\geq$2,000 \\
\midrule
T1 document type & $n$ & 0 & 17 & 111 & 314 & 590 & 455 & 214 & 75 \\
(TF-IDF + LR) & macro-F1 & -- & 0.958 & 0.918 & 0.902 & 0.923 & 0.884 & 0.833 & 0.839 \\
 & 95\% CI & -- & .81--1.0 & .85--.97 & .87--.94 & .90--.95 & .85--.92 & .77--.89 & .68--.95 \\
 & \cninety{} & -- & -- & 3 & 10 & 12.5 & 12.5 & 15 & 10 \\
\addlinespace
T2 dating & $n$ & 278 & 622 & 952 & 1,020 & 1,071 & 499 & 179 & 60 \\
(fastText) & MAE, years & 141 & 87 & 58 & 47 & 39 & 28 & 23 & 40 \\
 & 95\% CI & 122--161 & 76--97 & 52--65 & 42--53 & 36--43 & 24--32 & 18--29 & 21--65 \\
 & \cninety{} & 3 & 1 & 2 & 3 & 2 & 3 & 3 & 3 \\
\addlinespace
T3 doc.\ vs literary & $n$ & 484 & 908 & 1,183 & 1,214 & 1,262 & 593 & 213 & 93 \\
(TF-IDF + LR) & macro-F1 & 0.596 & 0.846 & 0.855 & 0.894 & 0.915 & 0.965 & 0.909 & 0.964 \\
 & 95\% CI & .49--.77 & .72--.94 & .79--.91 & .85--.93 & .87--.95 & .92--1.0 & .82--.97 & .90--1.0 \\
 & \cninety{} & (20) & 1 & 12.5 & 7.5 & 20 & 20 & 25 & 17.5 \\
\bottomrule
\end{tabular}
\end{table*}

\begin{table}[!tb]
\caption{\rev{Effective CER (\%) by document size at a nominal CER, default noise, test split (5,961 documents with letters; mean of the three noise seeds), and the share of documents that receive no edit. From 50 letters upward every document is edited at 2\% CER and above. The T1 papyri are degraded as a corpus of their own; all but 17 have 50 letters or more, and their effective CER is within 0.3 points of nominal at every level}}
\label{tab:effcer}
\centering\footnotesize\setlength{\tabcolsep}{3.5pt}\revcolor
\begin{tabular}{@{}rrrrrrr@{}}
\toprule
 & \multicolumn{2}{c}{1--19 letters} & \multicolumn{2}{c}{20--49 letters} & 50--99 & $\geq$100 \\
\cmidrule(lr){2-3}\cmidrule(lr){4-5}
Nominal & CER & no edit & CER & no edit & CER & CER \\
\midrule
1 & 0.00 & 100\% & 0.19 & 94\% & 1.35 & 1.00 \\
2 & 0.00 & 100\% & 2.44 & 18\% & 2.00 & 2.00 \\
3 & 1.71 & 80\% & 2.96 & 0\% & 3.12 & 3.00 \\
5 & 5.07 & 42\% & 4.73 & 0\% & 5.02 & 5.00 \\
10 & 10.68 & 10\% & 10.15 & 0\% & 10.07 & 9.99 \\
20 & 20.03 & 3\% & 20.00 & 0\% & 20.01 & 20.00 \\
30 & 30.54 & 1\% & 30.15 & 0\% & 30.09 & 29.99 \\
50 & 47.71 & 1\% & 49.26 & 0\% & 49.67 & 50.04 \\
\midrule
Documents & \multicolumn{2}{c}{489} & \multicolumn{2}{c}{908} & 1,184 & 3,380 \\
\bottomrule
\end{tabular}
\end{table}

\section{\revhead{Validation on real HTR output: protocol and further results}}\label{esm:protocol}
\movednote{Detail of Sect.~\ref{sec:real}.}

\rev{\textbf{Papyri.} The candidates are papyri with letters and an image link in the papyri.info metadata that Anagnostes has not met in any form (Sect.~\ref{sec:realset}). We selected 637 of them: the Grammateus papyri of the test split, which serve all four tasks; the literary texts of the test split; a sample of other dated test papyri, stratified by size; and Grammateus papyri of the other splits, for document type only. 372 reserve papyri of the same kinds stood in for papyri without an image. Of the 1,009 papyri attempted, 550 have no usable image: the holding collection forbids automated access or answers it with a challenge (among them Vienna and the British Library), or the link is dead. No image was fetched by hand. Of the 459 papyri read, 13 have no line that can be matched to the edition, and 73 of the remaining 446 fail the conditions of the final set (30 exceed 50\% CER, 40 have fewer than three matched lines and 50 fewer than 50 letters; many fail more than one).

The 373 papyri kept date from the fourth century BCE to the seventh CE; 155 come from the Oxyrhynchite and 105 from the Arsinoite nome; 362 are papyri, nine ostraca and two parchments. Their images come from 18 collections. Two sources supply 63\% of the papyri (the Oxyrhynchus collection of the Egypt Exploration Society, 132; photographs of papyri in Cairo, 104), followed by Michigan (36), Berlin (28), Cologne (20) and Florence (13). The 220 papyri with a Grammateus type are 89 of \emph{Epistolary Exchange}, 51 of \emph{Transmission of Information}, 46 of \emph{Objective Statement} and 34 of \emph{Recording of Information}. Of the 889 images, 695 carry lines that enter the comparison; the others are second photographs of the same text or show no line that was read. Document type is tested on 220 of the papyri (390 images, 4,273 matched lines, 127,871 letters, median CER 21.4\%), dating on 163 (465 images, 2,776 lines, 69,514 letters, 23.3\%), and documentary versus literary and search on 225 (617 images, 4,444 lines, 92,942 letters, 22.7\%).

\textbf{Recognition.} Lines are detected by the four-model line pipeline of Anagnostes (two detectors whose results are fused, with a refinement of the line ends). It had not yet replaced the detector of the public application when the validation was run, and none of the papyri is among its training pages. The lines are read by the recogniser of the application, a DINOv2 encoder with a TrOCR decoder, with a beam of width 8, without a language model and without the second stage of the system, which restores lost text and was trained on the editions. One setting serves every image, and no papyrus was read again because its result looked poor. Expansions of abbreviations, which the recogniser writes in round brackets, are dropped as they are in the corpus. When several photographs show the same text, the lines of the later ones are not scored.

\textbf{Alignment and CER.} Detected lines are aligned to the lines of the edition in order, one to one, by dynamic programming; a pair is accepted when the Levenshtein distance between the two lines is at most 60\% of the longer one. The threshold censors: a line read worse than this becomes a lost line instead of raising the CER. On the 446 scored papyri the pooled CER is 23.0\% at this threshold, 20.1\% at 50\% and 29.4\% at 75\%, where unrelated lines begin to be paired. As a check on chance pairs, every reading was also aligned to the edition of a different papyrus: about 2.5\% of its lines still find a partner. Five of the 373 papyri match no more lines than twice this level (four long literary rolls of which the image shows a small part, and one document); they are kept. Lost text is marked in the real output where the edition marks it, so that only letters differ between the versions. With the recogniser's own gap marks instead, the scores on real output move by at most 0.03 (macro-F1, F1 or recall) or 6.5 years.

\textbf{Twins.} For every papyrus the default noise of Sect.~\ref{sec:degrade} is applied to the clean L2 text of its matched lines with a budget of exactly the number of edits of the real output, under five seeds; every twin was checked to lie at that Levenshtein distance from the clean text. The default noise edits at most every second letter, so 20 papyri read with a CER above about 50\% have no twin; none of them is in the final set.

\textbf{Models.} The frozen models of Sect.~\ref{sec:models} are applied without further training. For T1 each papyrus is scored by the fold model that did not see it in training. For T4 the collection is the 4,444 matched lines of the 225 test papyri, and the queries are the 1,073 queries of Sect.~\ref{sec:t4data} that occur in their clean text. Of the search methods not shown in Table~\ref{tab:real}, the three filters that accept approximate hits are the only ones that do worse on real output than on the twins: within one edit (F1 0.111 against 0.140), within two edits (0.023 against 0.026) and within a normalised distance of 0.25 (0.183 against 0.197).

\textbf{Comparison.} Differences are tested by a paired percentile bootstrap with 1,000 draws of papyri (of queries for search), the same draw for the real output and for the five twins, whose scores are averaged. Table~\ref{tab:realbycer} gives the differences by measured CER.

\textbf{Error profile.} Table~\ref{tab:profile} compares the arrangement of the edits in the real output and in the twins (Sect.~\ref{sec:profile}). Real output differs in other ways as well. It drops more letters: 35\% of its edits are deletions, against 20\%. Its errors gather at the edges of lost text: the letters next to a gap, 4\% of all, receive 11\% of the substitutions and deletions. Its substitutions are more regular than chance but far from a small fixed set: the ten most frequent confusions, led by \gk{e} read as \gk{a} and \gk{o} as \gk{a}, account for 12\% of them, against 5\% in the twins. How the errors relate to the letters that editors mark as doubtful was not measured.

Of the variants of Sect.~\ref{sec:shape} the bursts come closest to this picture, but they do not reach it: real runs are often longer than two to four letters, and the document-type classifier keeps 93\% of its clean score on real output at a median CER of 28\%, where bursts leave it 80\% at 25\% CER and 58\% at 30\% on the test set (other papyri, so the comparison is only indicative). A noise model that stands in for real output would have to take its runs, its deletions and its place beside lost text from a profile like this one; we leave it to further work.

\textbf{Choices made after results had been seen.} Two choices followed the first results and are reported for that reason. The first run took its lines from the detector of the public application; the run reported here uses the newer line pipeline, with which the pooled CER of the papyri kept is 22.4\% instead of 25.5\%. The conditions on the papyri (a CER of at most 50\%, three matched lines, 50 letters) were set afterwards, because the default noise cannot produce a twin above about 50\% CER, because chance pairs sit near 58\% CER, and because a CER that rests on one or two short lines does not measure the recogniser. The main comparison was repeated without each choice, with one model per task as in Fig.~\ref{fig:real}. On the final set the differences between real output and twins are $+0.32$ macro-F1 for document type (TF-IDF), $-35$ years for dating (fastText), $+0.06$ macro-F1 for documentary versus literary (TF-IDF) and $+0.19$ F1 for search (exact match). On every scored papyrus, without the three conditions (234, 190 and 265 papyri for T1, T2 and T3), they are $+0.34$, $-39$, $+0.06$ and $+0.19$; with the final conditions and the first detector (215, 154 and 218 papyri) they are $+0.34$, $-42$, $+0.03$ and $+0.23$. Every 95\% interval excludes zero except that of documentary versus literary with the first detector. The differences for document type, dating and search thus keep their sign and size; the small difference for documentary versus literary does not hold with the first detector.}

\begin{table}[!ht]
\caption{\rev{How the 41,243 edits are arranged in the real HTR output of the 373 papyri and in their synthetic twins (mean of five), which carry the same number of edits. Letters beside lost text are 4\% of all letters}}
\label{tab:profile}
\centering\footnotesize\setlength{\tabcolsep}{3pt}\revcolor
\begin{tabular}{@{}>{\raggedright\arraybackslash}p{4.4cm}cc@{}}
\toprule
 & Real & Twins \\
\midrule
Substitutions : insertions : deletions (\%) & 49 : 16 : 35 & 60 : 20 : 20 \\
Edits that touch another edit & 75\% & 0\% \\
Edits in runs of five or more & 25\% & 0\% \\
Lines without an edit & 14\% & 6\% \\
Four-letter sequences of the edition kept & 60\% & 37\% \\
\quad at a CER above 20\% & 49\% & 20\% \\
Substitutions and deletions beside lost text & 11\% & 6\% \\
Substitutions in the ten most frequent confusions & 12\% & 5\% \\
\bottomrule
\end{tabular}
\end{table}

\begin{table}[!ht]
\caption{\rev{Real HTR output minus synthetic twins by measured CER of the papyrus: macro-F1 for T1 and T3, MAE in years for T2 (negative: real output is dated better), search over lines as in Table~\ref{tab:real}; the collection is the papyri of the group, so a query is counted in every group that contains it. An asterisk marks a 95\% interval that excludes zero}}
\label{tab:realbycer}
\centering\footnotesize\setlength{\tabcolsep}{3pt}\revcolor
\begin{tabular}{@{}lcccc@{}}
\toprule
Measured CER & 0--5\% & 5--10\% & 10--20\% & 20--50\% \\
\midrule
\multicolumn{5}{@{}l}{\textit{T1 document type}} \\
Papyri & 10 & 22 & 68 & 120 \\
TF-IDF & $-0.02$ & $+0.06$ & $+0.12$* & $+0.52$* \\
fastText & $-0.03$ & $+0.03$ & $+0.13$ & $+0.47$* \\
char-CNN & $+0.22$ & $+0.01$ & $+0.05$ & $+0.09$ \\
ByT5-small & $0.00$ & $+0.08$ & $+0.18$* & $+0.45$* \\
\addlinespace
\multicolumn{5}{@{}l}{\textit{T2 dating}} \\
Papyri & 7 & 15 & 37 & 104 \\
TF-IDF & $+1$ & $-10$ & $-9$ & $-30$* \\
fastText & $+4$ & $-1$ & $-24$* & $-47$* \\
char-CNN & $+3$ & $+4$ & $-16$* & $-18$* \\
ByT5-small & $+15$ & $-2$ & $-23$ & $-46$* \\
\addlinespace
\multicolumn{5}{@{}l}{\textit{T3 documentary versus literary}} \\
Papyri & 7 & 18 & 59 & 141 \\
TF-IDF & $0.00$ & $0.00$ & $+0.05$* & $+0.07$* \\
fastText & $0.00$ & $0.00$ & $+0.02$ & $0.00$ \\
char-CNN & $0.00$ & $+0.18$ & $-0.02$ & $-0.01$ \\
ByT5-small & $0.00$ & $-0.11$ & $+0.05$ & $+0.03$ \\
\addlinespace
\multicolumn{5}{@{}l}{\textit{T4 search}} \\
Queries & 163 & 347 & 643 & 908 \\
Exact match, F1 & $+0.05$* & $+0.07$* & $+0.13$* & $+0.26$* \\
BM25, R@20 & $+0.10$* & $+0.15$* & $+0.25$* & $+0.30$* \\
Trigram, R@20 & $+0.04$* & $+0.06$* & $+0.12$* & $+0.20$* \\
Min-edit, R@20 & $+0.01$ & $+0.02$* & $+0.04$* & $+0.09$* \\
\bottomrule
\end{tabular}
\end{table}

\end{document}

%% file: figures/fig1.tex
\providecommand{\rev}[1]{#1}
{\small\renewcommand{\arraystretch}{1.12}
\begin{tabular}{@{}p{54mm}r@{\hspace{2.5mm}}l@{}}
\textbf{Edition (L0)} & {19} & \gk{Kap'itwn[a] \d{p}\d{o}ll`a ka`i to\d{`u}\d{c} >adelfo'uc} \\
{\scriptsize as printed in the edition} & {20} & \gk{[m]ou ka`i Se[rhn'i]llan ka`i to[`uc] f'ilouc mo[u].} \\
\addlinespace[2pt]\midrule[0.3pt]\addlinespace[2pt]
\textbf{\rev{Preserved text (L1)}} & {19} & \gk{Kap'itwn\gap{} poll`a ka`i to`uc >adelfo'uc} \\
{\scriptsize editorial layer removed; \gap{} = lost text} & {20} & \gk{\gap{}ou ka`i Se\gap{}llan ka`i to\gap{} f'ilouc mo\gap{}.} \\
\addlinespace[2pt]\midrule[0.3pt]\addlinespace[2pt]
\textbf{Letters only \rev{(L2)}} & {19} & \gk{kapitwn\gap{}pollakaitousadelfous\textcompwordmark{}} \\
{\scriptsize = \rev{error-free} HTR output} & {20} & \gk{\gap{}oukaise\gap{}llankaito\gap{}filousmo\gap{}} \\
\addlinespace[2pt]\midrule[0.3pt]\addlinespace[2pt]
\textbf{5\% CER} & {19} & \gk{k\textbf{u}pit\textbf{s\textcompwordmark{}}n\gap{}pollakaitous\textcompwordmark{}adelfous\textcompwordmark{}} \\
{\scriptsize 3 edits in 51 letters} & {20} & \gk{\gap{}oukais\textcompwordmark{}e\gap{}lla\textbf{k}kaito\gap{}filous\textcompwordmark{}mo\gap{}} \\
\addlinespace[2pt]\midrule[0.3pt]\addlinespace[2pt]
\textbf{15\% CER} & {19} & \gk{kapitwn\gap{}po\textbf{u}lakaitou\textbf{$\cdot$}a\textbf{q}el\textbf{b}ous\textcompwordmark{}} \\
{\scriptsize 8 edits in 51 letters} & {20} & \gk{\gap{}oukais\textcompwordmark{}e\gap{}llan\textbf{w}a\textbf{u}to\gap{}fil\textbf{n}u\underline{\textbf{k}}s\textcompwordmark{}mo\gap{}} \\
\addlinespace[2pt]\midrule[0.3pt]\addlinespace[2pt]
\textbf{30\% CER} & {19} & \gk{kapitwn\gap{}\textbf{i}ol\underline{\textbf{e}}la\underline{\textbf{o}}kai\textbf{a}ou\underline{\textbf{w}}s\textcompwordmark{}a\textbf{$\cdot$}elf\underline{\textbf{m}}ou\underline{\textbf{o}}s\textcompwordmark{}} \\
{\scriptsize 15 edits in 51 letters} & {20} & \gk{\gap{}ouk\textbf{z}i\textbf{o}e\gap{}\textbf{d}la\textbf{a}ka\textbf{g}to\gap{}fi\underline{\textbf{u}}lou\underline{\textbf{x}}s\textcompwordmark{}mo\gap{}} \\
\end{tabular}}

%% file: figures/fig1_full.tex
\providecommand{\rev}[1]{#1}
{\small\renewcommand{\arraystretch}{1.12}
\begin{tabular}{@{}p{38mm}r@{\hspace{2.5mm}}l@{}}
\multirow[t]{5}{38mm}{\textbf{Edition (L0)}\\[1pt]{\scriptsize as printed: restorations, accents, case, word division}} & {1} & \gk{>Ap'iwn >Epim'aqw| t\~wi patr`i ka`i} \\
 & {2} & \gk{kur'iw| ple\~ista qa'irein. pr`o m`en p'an-} \\
 & {$\vdots$} &  \\
 & {19} & \gk{Kap'itwn[a] \d{p}\d{o}ll`a ka`i to\d{`u}\d{c} >adelfo'uc} \\
 & {20} & \gk{[m]ou ka`i Se[rhn'i]llan ka`i to[`uc] f'ilouc mo[u].} \\
\addlinespace[2pt]\midrule[0.3pt]\addlinespace[2pt]
\multirow[t]{5}{38mm}{\textbf{\rev{Preserved text (L1)}}\\[1pt]{\scriptsize editorial layer removed; \gap{} marks lost text}} & {1} & \gk{>Ap'iwn >Epim'aqw| t\~wi patr`i ka`i} \\
 & {2} & \gk{kur'iw| ple\~ista qa'irein. pr`o m`en p'an} \\
 & {$\vdots$} &  \\
 & {19} & \gk{Kap'itwn\gap{} poll`a ka`i to`uc >adelfo'uc} \\
 & {20} & \gk{\gap{}ou ka`i Se\gap{}llan ka`i to\gap{} f'ilouc mo\gap{}.} \\
\addlinespace[2pt]\midrule[0.3pt]\addlinespace[2pt]
\multirow[t]{5}{38mm}{\textbf{Letters only \rev{(L2)}}\\[1pt]{\scriptsize = \rev{error-free} HTR: lowercase, no diacritics, one sigma, no spaces}} & {1} & \gk{apiwnepimaqwtwipatrikai} \\
 & {2} & \gk{kuriwpleistaqaireinpromenpan} \\
 & {$\vdots$} &  \\
 & {19} & \gk{kapitwn\gap{}pollakaitousadelfous\textcompwordmark{}} \\
 & {20} & \gk{\gap{}oukaise\gap{}llankaito\gap{}filousmo\gap{}} \\
\addlinespace[2pt]\midrule[0.3pt]\addlinespace[2pt]
\multirow[t]{5}{38mm}{\textbf{5\% CER}\\[1pt]{\scriptsize 5 edits in 102 letters}} & {1} & \gk{a\textbf{l}iwnepimaqwtwipatrikai} \\
 & {2} & \gk{kuriwpleis\textcompwordmark{}taqa\textbf{z}reinpromenpan} \\
 & {$\vdots$} &  \\
 & {19} & \gk{kapi\underline{\textbf{j}}tw\textbf{z}\gap{}pollakaitous\textcompwordmark{}adelfo\textbf{q}s\textcompwordmark{}} \\
 & {20} & \gk{\gap{}oukais\textcompwordmark{}e\gap{}llankaito\gap{}filous\textcompwordmark{}mo\gap{}} \\
\addlinespace[2pt]\midrule[0.3pt]\addlinespace[2pt]
\multirow[t]{5}{38mm}{\textbf{15\% CER}\\[1pt]{\scriptsize 15 edits in 102 letters}} & {1} & \gk{ap\textbf{j}wnepima\textbf{h}w\textbf{u}wipa\textbf{s\textcompwordmark{}}rikai} \\
 & {2} & \gk{k\textbf{j}riwpleis\textcompwordmark{}t\textbf{k}qaireinp\underline{\textbf{a}}ro\textbf{o}enpan} \\
 & {$\vdots$} &  \\
 & {19} & \gk{kapit\textbf{$\cdot$}n\gap{}p\textbf{u}ll\textbf{$\cdot$}k\textbf{$\cdot$}ito\textbf{$\cdot$}s\textcompwordmark{}ad\underline{\textbf{h}}elfous\textcompwordmark{}} \\
 & {20} & \gk{\gap{}ouka\underline{\textbf{\textstigma{}}}is\textcompwordmark{}e\gap{}llankaito\gap{}filous\textcompwordmark{}mo\gap{}} \\
\addlinespace[2pt]\midrule[0.3pt]\addlinespace[2pt]
\multirow[t]{5}{38mm}{\textbf{30\% CER}\\[1pt]{\scriptsize 31 edits in 102 letters}} & {1} & \gk{\textbf{s\textcompwordmark{}}piw\textbf{s\textcompwordmark{}}epi\underline{\textbf{\textqoppa{}}}ma\underline{\textbf{u}}qw\underline{\textbf{z}}tw\textbf{b}p\textbf{\textsampi{}}t\underline{\textbf{d}}rik\textbf{f}i} \\
 & {2} & \gk{kuriw\textbf{w}le\textbf{m}s\textcompwordmark{}taqair\textbf{i}i\underline{\textbf{l}}npromenpa\textbf{$\cdot$}} \\
 & {$\vdots$} &  \\
 & {19} & \gk{\textbf{t}apit\underline{\textbf{d}}wn\gap{}\textbf{f}o\textbf{t}laka\underline{\textbf{s\textcompwordmark{}}}it\textbf{d}us\textcompwordmark{}\textbf{t}del\textbf{$\cdot$}o\textbf{$\cdot$}s\textcompwordmark{}} \\
 & {20} & \gk{\gap{}\textbf{y}u\textbf{y}ais\textcompwordmark{}\textbf{i}\gap{}\underline{\textbf{h}}llankai\textbf{i}o\gap{}\underline{\textbf{g}}filo\textbf{a}s\textcompwordmark{}\underline{\textbf{h}}mo\gap{}} \\
\end{tabular}}